\documentclass[twoside]{article}
\usepackage{jair,theapa,rawfonts}

\usepackage{varioref}
\usepackage{graphicx}
\usepackage{url}

\def\mytitle{Algorithm Selection for Combinatorial Search Problems:\\A survey}
\def\myauthor{Lars Kotthoff}

\title{\mytitle}
\author{\name \myauthor \email larsko@4c.ucc.ie}

\begin{document}

\ShortHeadings{Algorithm Selection for Search: A survey}{Kotthoff}

\maketitle

\begin{abstract} 
The Algorithm Selection Problem is concerned with selecting the best algorithm
to solve a given problem on a case-by-case basis. It has become especially
relevant in the last decade, as researchers are increasingly investigating how
to identify the most suitable existing algorithm for solving a problem instead
of developing new algorithms. This survey presents an overview of this work
focusing on the contributions made in the area of combinatorial search problems,
where Algorithm Selection techniques have achieved significant performance
improvements. We unify and organise the vast literature according to criteria
that determine Algorithm Selection systems in practice. The comprehensive
classification of approaches identifies and analyses the different directions
from which Algorithm Selection has been approached. This paper contrasts and
compares different methods for solving the problem as well as ways of using
these solutions. It closes by identifying directions of current and future
research.
\end{abstract}

\section{Introduction}

For many years, Artificial Intelligence research has been focusing on inventing
new algorithms and approaches for solving similar kinds of problems. In some
scenarios, a new algorithm is clearly superior to previous approaches. In the
majority of cases however, a new approach will improve over the current state of
the art only for some problems. This may be because it employs a heuristic that
fails for problems of a certain type or because it makes other assumptions about
the problem or environment that are not satisfied in some cases. Selecting the
most suitable algorithm for a particular problem aims at mitigating these
problems and has the potential to significantly increase performance in
practice. This is known as the Algorithm Selection Problem.

The Algorithm Selection Problem has, in many forms and with different names,
cropped up in many areas of Artificial Intelligence in the last few decades.
Today there exists a large amount of literature on it. Most publications are
concerned with new ways of tackling this problem and solving it efficiently in
practice. Especially for combinatorial search problems, the application of
Algorithm Selection techniques has resulted in significant performance
improvements that leverage the diversity of systems and techniques developed in
recent years. This paper surveys the available literature and describes how
research has progressed.

Researchers have long ago recognised that a single algorithm will not give the
best performance across all problems one may want to solve and that selecting
the most appropriate method is likely to improve the overall performance.
Empirical evaluations have provided compelling evidence for this
\cite<e.g.>{aha_generalizing_1992,wolpert_no_1997}.

\begin{figure}[tp]
\begin{center}
\includegraphics[width=\textwidth]{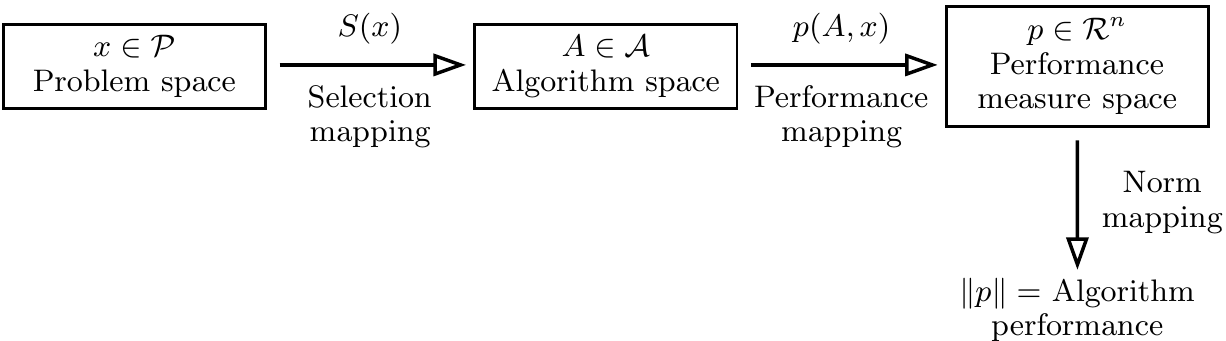}
\end{center}
\caption{Basic model for the Algorithm Selection Problem
as published in \protect\citeA{rice_algorithm_1976}.}
\label{algselectionorig}
\end{figure}

The original description of the Algorithm Selection Problem was published in
\citeA{rice_algorithm_1976}. The basic model described in the paper is very
simple -- given a space of problems and a space of algorithms, map each
problem-algorithm pair to its performance. This mapping can then be used to
select the best algorithm for a given problem. The original figure that
illustrates the model is reproduced in Figure~\vref{algselectionorig}. As Rice
states,
\begin{quote}
``The objective is to determine $S(x)$ [the mapping of problems to algorithms]
so as to have high algorithm performance.''
\end{quote}

He identifies the following four criteria for the selection process.
\begin{enumerate}
\item Best selection for all mappings $S(x)$ and problems $x$. For every
    problem, an algorithm is chosen to give maximum performance.
\item Best selection for a subclass of problems. A single algorithm is chosen to
    apply to each of a subclass of problems such that the performance
    degradation compared to choosing from all algorithms is minimised.
\item Best selection from a subclass of mappings. Choose the selection mapping
    from a subset of all mappings from problems to algorithms such that the
    performance degradation is minimised.
\item Best selection from a subclass of mappings and problems. Choose a single
    algorithm from a subset of all algorithms to apply to each of a subclass of
    problems such that the performance degradation is minimised.
\end{enumerate}
The first case is clearly the most desirable one. In practice however, the other
cases are more common -- we might not have enough data about individual problems
or algorithms to select the best mapping for everything.

\citeA{rice_algorithm_1976} lists five main steps for solving the problem.
\begin{description}
\item[Formulation] Determination of the subclasses of problems and mappings to
    be used.
\item[Existence] Does a best selection mapping exist?
\item[Uniqueness] Is there a unique best selection mapping?
\item[Characterization] What properties characterize the best selection mapping
    and serve to identify it?
\item[Computation] What methods can be used to actually obtain the best
    selection mapping?
\end{description}\label{solframework}
This framework is taken from the theory of approximation of functions. The
questions for existence and uniqueness of a best selection mapping are usually
irrelevant in practice. As long as a \emph{good} performance mapping is found
and improves upon the current state of the art, the question of whether there is
a different mapping with the same performance or an even better mapping is
secondary. While it is easy to determine the theoretically best selection
mapping on a set of given problems, casting this mapping into a
\emph{generalisable} form that will give good performance on new problems or
even into a form that can be used in practice is hard. Indeed,
\citeA{guo_algorithm_2003} shows that the Algorithm Selection Problem in general
is undecidable. It may be better to choose a mapping that generalises well
rather than the one with the best performance. Other considerations can be
involved as well. \citeA{guo_learning-based_2004} and
\citeA{cook_maximizing_1997} compare different Algorithm selection models and
select not the one with the best performance, but one with good performance that
is also easy to understand, for example. \citeA{vrakas_learning_2003} select
their method of choice for the same reason. Similarly, \citeA{xu_satzilla_2008}
choose a model that is cheap to compute instead of the one with the best
performance. They note that,
\begin{quote}
``All of these techniques are computationally more expensive than ridge
regression, and in our previous experiments we found that they did not improve
predictive performance enough to justify this additional cost.''
\end{quote}

\begin{figure}[tp]
\begin{center}
\includegraphics[width=\textwidth]{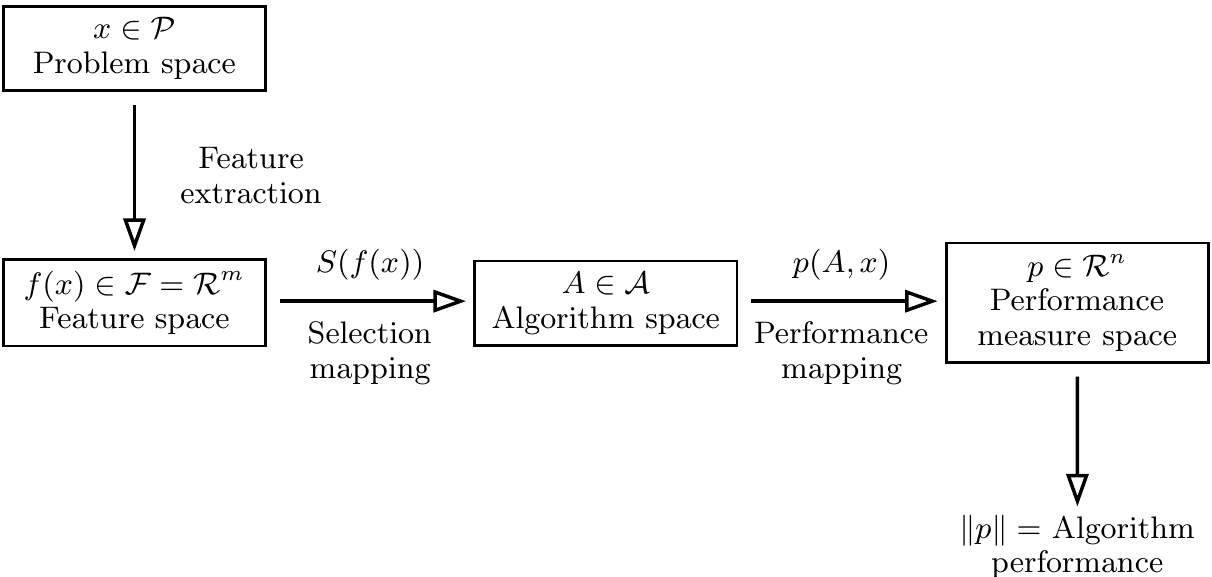}
\end{center}
\caption{Refined model for the Algorithm Selection Problem with problem
features \protect\cite{rice_algorithm_1976}.}
\label{algselectionfeaturesorig}
\end{figure}

Rice continues by giving practical examples of where his model applies. He
refines the original model to include features of problems that can be used to
identify the selection mapping. The original figure depicting the refined model
is given in Figure~\ref{algselectionfeaturesorig}. This model, or a variant of
it, is what is used in most practical approaches. Including problem features is
the crucial difference that often makes an approach feasible.

For each problem in a given set, the features are extracted. The aim is to use
these features to produce the mapping that selects the algorithm with the best
performance for each problem. The actual performance mapping for each
problem-algorithm pair is usually of less interest as long as the individual
best algorithm can be identified.

Rice poses additional questions about the determination of features.
\begin{itemize}
\item What are the best features for predicting the performance of a specific
    algorithm?
\item What are the best features for predicting the performance of a specific
    class of algorithms?
\item What are the best features for predicting the performance of a subclass of
    selection mappings?
\end{itemize}
He also states that,
\begin{quote}
``The determination of the best (or even good) features is one of the most
important, yet nebulous, aspects of the algorithm selection problem.''
\end{quote}
He refers to the difficulty of knowing the problem space. Many problem spaces
are not well known and often a sample of problems is drawn from them to
evaluate empirically the performance of the given set of algorithms. If the
sample is not representative, or the features do not facilitate a good
separation of the problem classes in the feature space, there is little hope of
finding the best or even a good selection mapping.

\citeA{vassilevska_confronting_2006} note that,
\begin{quote}
``While it seems that restricting a heuristic to a special case would likely
improve its performance, we feel that the ability to partition the problem
space of some $\mathcal{NP}$-hard problems by efficient selectors is mildly
surprising.''
\end{quote}
This sentiment was shared by many researchers and part of the great prominence
of Algorithm Selection systems especially for combinatorial search problems can
probably be attributed to the surprise that it actually works.

Most approaches employ Machine Learning to learn the performance mapping from
problems to algorithms using features extracted from the problems. This often
involves a \emph{training phase}, where the candidate algorithms are run on a
sample of the problem space to experimentally evaluate their performance. This
training data is used to create a \emph{performance model} that can be used to
predict the performance on new, unseen problems. The term \emph{model} is used
only in the loosest sense here; it can be as simple as a representation of the
training data without any further analysis.

\subsection{Practical motivation}

\citeA{aha_generalizing_1992} notes that in Machine Learning, researchers often
perform experiments on a limited number of data sets to demonstrate the
performance improvements achieved and implicitly assume that these improvements
generalise to other data. He proposes a framework for better experimental
evaluation of such claims and deriving rules that determine the properties a
data set must have in order for an algorithm to have superior performance. His
objective is
\begin{quote}
``\ldots to derive rules of the form `this algorithm outperforms these other
algorithms on these dependent measures for databases with these
characteristics'. Such rules summarize \emph{when} [\ldots] rather than
\emph{why} the observed performance difference occurred.''
\end{quote}

\citeA{tsang_attempt_1995} make similar observations and show that there is no
algorithm that is universally the best when solving constraint problems. They
also demonstrate that the best algorithm-heuristic combination is not what one
might expect for some of the surveyed problems. This provides an important
motivation for research into performing Algorithm Selection automatically. They
close by noting that,
\begin{quote}
``\ldots research should focus on how to retrieve the most efficient
[algorithm-heuristic] combinations for a problem.''
\end{quote}

The focus of Algorithm Selection is on identifying algorithms with good
performance, not on providing explanations for why this is the case. Most
publications do not consider the question of ``Why?'' at all. Rice's framework
does not address this question either. The simple reason for this is that
explaining the Why? is difficult and for most practical applications not
particularly relevant as long as improvements can be achieved. Research into
what makes a problem hard, how this affects the behaviour of specific algorithms
and how to exploit this knowledge is a fruitful area, but outside the scope of
this paper. However, we present a brief exposition of one of the most important
concepts to illustrate its relevance.

The notion of a \emph{phase transition} \cite{cheeseman_where_1991} refers to a
sudden change in the hardness of a problem as the value of a single parameter of
the problem is changed. Detecting such transitions is an obvious way to
facilitate Algorithm Selection. \citeA{hogg_phase_1996} note that,
\begin{quote}
``In particular, the location of the phase transition point might provide a
systematic basis for selecting the type of algorithm to use on a given
problem.''
\end{quote}
While some approaches make use of this knowledge to generate challenging
training problems for their systems, it is hardly used at all to facilitate
Algorithm Selection. \citeA{nudelman_understanding_2004} use a set of features
that can be used to characterise a phase transition and note that,
\begin{quote}
``It turns out that [\ldots] this group of features alone suffices to construct
reasonably good models.''
\end{quote}
It remains unclear how relevant phase transitions are to Algorithm Selection in
practice. On one hand, their theoretical properties seem to make them highly
suitable, but on the other hand almost nobody has explored their use in actual
Algorithm Selection systems.

\subsubsection{No Free Lunch theorems}

The question arises of whether, in general, the performance of a system can be
improved by always picking the best algorithm. The ``No Free Lunch'' (NFL)
theorems \cite{wolpert_no_1997} state that no algorithm can be the best across
all possible problems and that on average, all algorithms perform the same. This
seems to provide a strong motivation for Algorithm Selection -- if, on average,
different algorithms are the best for different parts of the problem space,
selecting them based on the problem to solve has the potential to improve
performance.

The theorems would apply to Algorithm Selection systems themselves as well
though (in particular the version for supervised learning are
relevant, see \citeR{wolpert_supervised_2001}). This means that although
performance improvements can be achieved by selecting the right algorithms on
one part of the problem space, wrong decisions will be made on other parts,
leading to a loss of performance. On average over all problems, the performance
achieved by an Algorithm Selection meta-algorithm will be the same as that of
all other algorithms.

The NFL theorems are the source of some controversy however. Among the
researchers to doubt their applicability is the first proponent of the Algorithm
Selection Problem \cite{rice_how_1999}. Several other publications show that
the assumptions underlying the NFL may not be
satisfied \cite{rao_for_1995,domingos_how_1998}. In particular, the distribution
of the best algorithms from the portfolio to problems is not random -- it is
certainly true that certain algorithms are the best on a much larger number of
problems than others.

A detailed assessment of the applicability of the NFL theorems to the Algorithm
Selection Problem is outside the scope of this paper. However, a review of the
literature suggests that, if the theorems are applicable, the ramifications in
practice may not be significant. Most of the many publications surveyed here do
achieve performance improvements across a range of different problems using
Algorithm Selection techniques. As a research area, it is very active and
thriving despite the potentially negative implications of the NFL.

\subsection{Scope and related work}

Algorithm Selection is a very general concept that applies not only in almost
all areas of Computer Science, but also other disciplines. However, it is
especially relevant in many areas of Artificial Intelligence. This is a large
field itself though and surveying all Artificial Intelligence publications that
are relevant to Algorithm Selection in a single paper is infeasible.

In this paper, we focus on Algorithm Selection for \emph{combinatorial search
problems}. This is a large and important subfield of Artificial Intelligence
where Algorithm Selection techniques have become particularly prominent in
recent years because of the impressive performance improvements that have been
achieved by some approaches. Combinatorial search problems include for example
satisfiability (SAT), constraint problems, planning, quantified Boolean formulae
(QBF), scheduling and combinatorial optimisation.

A combinatorial search problem is one where an initial state is to be
transformed into a goal state by application of a series of operators, such as
assignment of values to variables. The space of possible states is usually
exponential in the size of the input and finding a solution is
$\mathcal{NP}$-hard. A common way of solving such problems is to use
\emph{heuristics}. A heuristic is a strategy that determines which operators to
apply when. Heuristics are not necessarily complete or deterministic, i.e.\ they
are not guaranteed to find a solution if it exists or to always make the same
decision under the same circumstances. The nature of heuristics makes them
particularly amenable to Algorithm Selection -- choosing a heuristic manually is
difficult even for experts, but choosing the correct one can improve performance
significantly.

Several doctoral dissertations with related work chapters that survey the
literature on Algorithm Selection have been produced. Examples of the more
recent ones include
\citeA{streeter_using_2007,hutter_automated_2009,carchrae_low_2009,gagliolo_online_2010,ewald_automatic_2010,kotthoff_algorithm_2012,malitsky_thesis_2012}.
\citeA{smith-miles_cross-disciplinary_2009} presents a survey with similar aims.
It looks at the Algorithm Selection Problem from the Machine Learning point of
view and focuses on seeing Algorithm Selection as a learning problem. As a
consequence, great detail is given for aspects that are relevant to Machine
Learning. In this paper, we take a more practical point of view and focus on
techniques that facilitate and implement Algorithm Selection systems. We are
furthermore able to take more recent work in this fast-moving area into account.

In contrast to most other work surveying Algorithm Selection literature, we take
an approach-centric view instead of a literature-centric one. This means that
instead of analysing a particular publication or system according to various
criteria, the different aspects of Algorithm Selection are illustrated with
appropriate references. A single publication may therefore appear in different
sections of this paper, giving details on different aspects of the authors'
approach.

There exists a large body of work that is relevant to Algorithm Selection in the
Machine Learning literature. \citeA{smith-miles_cross-disciplinary_2009}
presents a survey of many approaches. Repeating this here is unnecessary and
outside the scope of this paper, which focuses on the application of such
techniques. The most relevant area of research is that into \emph{ensembles},
where several models are created instead of one. Such ensembles are either
implicitly assumed or explicitly engineered so that they complement each other.
Errors made by one model are corrected by another. Ensembles can be engineered
by techniques such as \emph{bagging} \cite{breiman_bagging_1996} and
\emph{boosting} \cite{schapire_strength_1990}.
\citeA{bauer_empirical_1999,opitz_popular_1999} present studies that compare
bagging and boosting empirically. \citeA{dietterich_ensemble_2000} provides
explanations for why ensembles can perform better than individual algorithms.

There is increasing interest in the integration of Algorithm Selection
techniques with programming language paradigms
\cite<e.g.>{ansel_petabricks_2009,hoos_programming_2012}. While these issues are
sufficiently relevant to be mentioned here, exploring them in detail is outside
the scope of the paper. Similarly, technical issues arising from the
computation, storage and application of performance models, the integration of
Algorithm Selection techniques into complex systems, the execution of choices
and the collection of experimental data to facilitate Algorithm Selection are
not surveyed here.

\subsection{Terminology}

Algorithm Selection is a widely applicable concept and as such has cropped up
frequently in various lines of research. Often, different terminologies are
used.

\citeA{borrett_adaptive_1996} use the term \emph{algorithm chaining} to mean
switching from one algorithm to another while the problem is being solved.
%This is an instance of online Algorithm Selection.
\citeA{lobjois_branch_1998} call
Algorithm Selection \emph{selection by performance prediction}.
\citeA{vassilevska_confronting_2006} use the term \emph{hybrid algorithm} for
the combination of a set of algorithms and an Algorithm Selection model (which
they term \emph{selector}).

In Machine Learning, Algorithm Selection is usually referred to as
\emph{meta-learning}. This is because Algorithm Selection models for Machine
Learning learn when to use which method of Machine Learning. The earliest
approaches also spoke of \emph{hybrid approaches}
\cite<e.g.>{utgoff_perceptron_1988}. \citeA{aha_generalizing_1992} proposes
rules for selecting a Machine Learning algorithm that take the characteristics
of a data set into account. He uses the term \emph{meta-learning}.
\citeA{brodley_automatic_1993} introduces the notion of \emph{selective
superiority}. This concept refers to a particular algorithm being best on some,
but not all tasks.

%In heuristics research, the terms used for Algorithm Selection are
%\emph{meta-heuristic} or \emph{hyper-heuristic}. An Algorithm Selection model
%can be seen as a heuristic that decides when to use one of a set of heuristics.
%The term \emph{hyper-heuristic} was first used by
%\citeA{cowling_parameter-free_2001}. The term \emph{meta-heuristic} is part of
%Artificial Intelligence folklore and it is hard to trace its exact origins. The
%first mention of the term was probably in the paper that proposed Tabu search
%\cite{glover_future_1986}. It should be noted that these terms are not
%necessarily synonymous with Algorithm Selection, but may denote a superset of
%research in some areas. Hyper-heuristics research for example is also concerned
%with the development of new heuristics in some areas.

In addition to the many terms used for the process of Algorithm Selection,
researchers have also used different terminology for the models of what Rice
calls \emph{performance measure space}. \citeA{allen_selecting_1996} call them
\emph{runtime performance predictors}.
\citeA{leyton-brown_learning_2002,hutter_performance_2006,xu_hierarchical_2007,leyton-brown_empirical_2009}
coined the term \emph{Empirical Hardness model}. This stresses the reliance on
empirical data to create these models and introduces the notion of
\emph{hardness} of a problem. The concept of hardness takes into account all
performance considerations and does not restrict itself to, for example, runtime
performance. In practice however, the described empirical hardness models only
take runtime performance into account. In all cases, the predicted measures are
used to select an algorithm.

\medskip

Throughout this paper, the term \emph{algorithm} is used to refer to what is
selected for solving a problem. This is for consistency and to make the
connection to Rice's framework. An algorithm may be a system, a programme, a
heuristic, a classifier or a configuration. This is not made explicit unless it
is relevant in the particular context.

\subsection{Organisation}

An organisation of the Algorithm Selection literature is challenging, as there
are many different criteria that can be used to classify it. Each publication
can be evaluated from different points of view. The organisation of this paper
follows the main criteria below.

\begin{description}
\item[What to select algorithms from]\hfill\\
    Section~\ref{sec:portfolios} describes how sets of algorithms, or
    \emph{portfolios}, can be constructed. A portfolio can be \emph{static},
    where the designer decides which algorithms to include, or \emph{dynamic},
    where the composition or individual algorithms vary or change for different
    problems.
\item[What to select and when]\hfill\\
    Section~\ref{sec:solving} describes how algorithms from portfolios are
    selected to solve problems. Apart from the obvious approach of picking a
    single algorithm, time slots can be allocated to all or part of the
    algorithms or the execution monitored and earlier decisions revised.
    We also distinguish between selecting before the solving of the actual
    problem starts and while the problem is being solved.
\item[How to select]\hfill\\
    Section~\ref{sec:selectors} surveys techniques used for making the choices
    described in Section~\ref{sec:solving}. It details how performance models
    can be built and what kinds of predictions they inform. Example predictions
    are the best algorithm in the portfolio and the runtime performance of each
    portfolio algorithm.
\item[How to facilitate the selection]\hfill\\
    Section~\ref{sec:features} gives an overview of the types of analysis
    different approaches perform and what kind of information is gathered to
    facilitate Algorithm Selection. This includes the past performance of
    algorithms and structural features of the problems to be solved.
\end{description}

The order of the material follows a top-down approach. Starting with the
high-level idea of Algorithm Selection, as proposed by
\citeA{rice_algorithm_1976} and described in this introduction, more technical
details are gradually explored. Earlier concepts provide motivation and context
for later technical details. For example, the choice of whether to select a
single algorithm or monitor its execution (Section~\ref{sec:solving}) determines
the types of predictions required and techniques suitable for making them
(Section~\ref{sec:selectors}) as well as the properties that need to be measured
(Section~\ref{sec:features}).

The individual sections are largely self-contained. If the reader is more
interested in a bottom-up approach that starts with technical details on what
can be observed and measured to facilitate Algorithm Selection,
Sections~\ref{sec:portfolios} through~\ref{sec:features} may be read in reverse
order.

Section~\ref{sec:domains} again illustrates the importance of the field by
surveying the many different application domains of Algorithm Selection
techniques with a focus on combinatorial search problems. We close by briefly
discussing current and future research directions in
Section~\ref{sec:directions} and summarising in Section~\ref{sec:conclusion}.

%\begin{figure}[tp]
%\begin{center}
%\includegraphics[width=\textwidth]{taxonomy}
%\end{center}
%\caption{Organisation.}
%\label{org}
%\end{figure}

\section{Algorithm portfolios}\label{sec:portfolios}

For diverse sets of problems, it is unlikely that a single algorithm will be the
most suitable one in all cases. A way of mitigating this restriction is to use
a \emph{portfolio} of algorithms. This idea is closely related to the notion of
Algorithm Selection itself -- instead of making an up-front decision on what
algorithm to use, it is decided on a case-by-case basis for each problem
individually. In the framework presented by \citeA{rice_algorithm_1976},
portfolios correspond to the algorithm space $\mathcal{A}$.

Portfolios are a well-established technique in Economics. Portfolios of assets,
securities or similar products are used to reduce the risk compared to holding
only a single product. The idea is simple -- if the value of a single security
decreases, the total loss is less severe. The problem of allocating funds to the
different parts of the portfolio is similar to allocating resources to
algorithms in order to solve a computational problem. There are some important
differences though. Most significantly, the past performance of an algorithm can
be a good indicator of future performance. There are fewer factors that affect
the outcome and in most cases, they can be measured directly. In Machine
Learning, \emph{ensembles} \cite{dietterich_ensemble_2000} are instances of
algorithm portfolios. In fact, the only difference between algorithm portfolios
and Machine Learning ensembles is the way in which its constituents are used.

The idea of algorithm portfolios was first presented by
\citeA{huberman_economics_1997}. They describe a formal framework for the
construction and application of algorithm portfolios and evaluate their approach
on graph colouring problems. Within the Artificial Intelligence community,
algorithm portfolios were popularised by
\citeA{gomes_algorithm_1997,gomes_practical_1997} and a subsequent extended
investigation \cite{gomes_algorithm_2001}. The technique itself however had
been described under different names by other authors at about the same time in
different contexts.

\citeA{tsang_attempt_1995} experimentally show for a selection of constraint
satisfaction algorithms and heuristics that none is the best on all evaluated
problems. They do not mention portfolios, but propose that future research
should focus on identifying when particular algorithms and heuristics deliver
the best performance. This implicitly assumes a portfolio to choose algorithms
from. \citeA{allen_selecting_1996} perform a similar investigation and come to
similar conclusions. They talk about selecting an appropriate algorithm from an
\emph{algorithm family}.

Beyond the simple idea of using a set of algorithms instead of a single one,
there is a lot of scope for different approaches. One of the first problems
faced by researchers is how to construct the portfolio. There are two main
types. \emph{Static portfolios} are constructed offline before any problems are
solved. While solving a problem, the composition of the portfolio and the
algorithms within it do not change. \emph{Dynamic portfolios} change in
composition, configuration of the constituent algorithms or both during solving.

\subsection{Static portfolios}

Static portfolios are the most common type. The number of algorithms or systems
in the portfolio is fixed, as well as their parameters. In Rice's notation, the
algorithm space $\mathcal{A}$ is constant, finite and known. This approach is
used for example in SATzilla
\cite{nudelman_understanding_2004,xu_satzilla-07_2007,xu_satzilla_2008}, AQME
\cite{pulina_multi-engine_2007,pulina_self-adaptive_2009}, CPhydra
\cite{omahony_using_2008}, \textsc{ArgoSmArT}
\cite{nikoli_instance-based_2009} and BUS \cite{howe_exploiting_1999}.

The vast majority of approaches composes static portfolios from different
algorithms or different algorithm configurations.
\citeA{huberman_economics_1997} however use a portfolio that contains the same
randomised algorithm twice. They run the portfolio in parallel and as such
essentially use the technique to parallelise an existing sequential algorithm.

Some approaches use a large number of algorithms in the portfolio, such as
ArgoSmArT, whose portfolio size is 60. SATzilla uses 19 algorithms, although the
authors use portfolios containing only subsets of those for specific
applications. BUS uses six algorithms and CPhydra five.
\citeA{gent_learning_2010} select from a portfolio of only two algorithms. AQME
has different versions with different portfolio sizes, one with 16 algorithms,
one with five and three algorithms of different types and one with two
algorithms \cite{pulina_self-adaptive_2009}. The authors compare the different
portfolios and conclude that the one with eight algorithms offers the best
performance, as it has more variety than the portfolio with two algorithms and
it is easier to make a choice for eight than for 16 algorithms. There are also
approaches that use portfolios of variable size that is determined by training
data \cite{kadioglu_isac_2010,xu_hydra_2010}.

As the algorithms in the portfolio do not change, their selection is crucial for
its success. Ideally, the algorithms will complement each other such that good
performance can be achieved on a wide range of different problems.
\citeA{hong_groups_2004} report that portfolios composed of a random selection
from a large pool of diverse algorithms outperform portfolios composed of the
algorithms with the best overall performance. They develop a framework with a
mathematical model that theoretically justifies this observation.
\citeA{samulowitz_learning_2007} use a portfolio of heuristics for solving
quantified Boolean formulae problems that have specifically been crafted to be
orthogonal to each other. \citeA{xu_hydra_2010} automatically engineer a
portfolio with algorithms of complementary strengths. In
\citeA{xu_evaluating_2012}, the authors analyse the contributions of the
portfolio constituents to the overall performance and conclude that not
algorithms with the best overall performance, but with techniques that set them
apart from the rest contribute most. \citeA{kadioglu_isac_2010} use a static
portfolio of variable size that adapts itself to the training data. They cluster
the training problems and choose the best algorithm for each cluster. They do
not emphasise diversity, but suitability for distinct parts of the problem
space. \citeA{xu_hydra_2010} also construct a portfolio with algorithms that
perform well on different parts of the problem space, but do not use clustering.

In financial theory, constructing portfolios can be seen as a quadratic
optimisation problem. The aim is to balance expected performance and risk (the
expected variation of performance) such that performance is maximised and risk
minimised. \citeA{ewald_selecting_2010} solve this problem for algorithm
portfolios using genetic algorithms.

Most approaches make the composition of the portfolio less explicit. Many
systems use portfolios of solvers that have performed well in solver
competitions with the implicit assumption that they have complementing strengths
and weaknesses and the resulting portfolio will be able to achieve good
performance.

\subsection{Dynamic portfolios}

Rather than relying on a priori properties of the algorithms in the portfolio,
dynamic portfolios adapt the composition of the portfolio or the algorithms
depending on the problem to be solved. The algorithm space $\mathcal{A}$ changes
with each problem and is a subspace of the potentially infinite super algorithm
space $\mathcal{A}'$. This space contains all possible (hypothetical) algorithms
that could be used to solve problems from the problem space. In static
portfolios, the algorithms in the portfolio are selected from $\mathcal{A}'$
once either manually by the designer of the portfolio or automatically based on
empirical results from training data.

One approach is to build a portfolio by combining algorithmic building blocks.
An example of this is the Adaptive Constraint Engine (ACE)
\cite{epstein_collaborative_2001,epstein_adaptive_2002}. The building blocks are
so-called advisors, which characterise variables of the constraint problem and
give recommendations as to which one to process next. ACE combines these
advisors into more complex ones.
\citeA{elsayed_synthesis_2010,elsayed_synthesis_2011} use a similar idea to
construct search strategies for solving constraint problems.
\citeA{fukunaga_automated_2002,fukunaga_automated_2008} proposes CLASS, which
combines heuristic building blocks to form composite heuristics for solving SAT
problems. In these approaches, there is no strong notion of a portfolio -- the
algorithm or strategy used to solve a problem is assembled from lower level
components.

Closely related is the concept of specialising generic building blocks for the
problem to solve. This approach is taken in the SAGE system (Strategy
Acquisition Governed by Experimentation)
\cite{langley_learningd_1983,langley_learning_1983}. It starts with a set of
general operators that can be applied to a search state. These operators are
refined by making the preconditions more specific based on their utility for
finding a solution. The \textsc{Multi-tac} (Multi-tactic Analytic Compiler)
system
\cite{minton_integrating_1993,minton_analytic_1993,minton_automatically_1996}
specialises a set of generic heuristics for the constraint problem to solve.

There can be complex restrictions on how the building blocks are combined.
RT-Syn \cite{smith_knowledge-based_1992} for example uses a preprocessing step
to determine the possible combinations of algorithms and data structures to
solve a software specification problem and then selects the most appropriate
combination using simulated annealing. \citeA{balasubramaniam_automated_2012}
model the construction of a constraint solver from components as a constraint
problem whose solutions denote valid combinations of components.

Another approach is to modify the parameters of parameterised algorithms in the
portfolio. This is usually referred to as automatic tuning and not only
applicable in the context of algorithm portfolios, but also for single
algorithms. The HAP system \cite{vrakas_learning_2003} automatically tunes
the parameters of a planning system depending on the problem to solve.
\citeA{horvitz_bayesian_2001} dynamically modify algorithm parameters during
search based on statistics collected during the solving process.

\subsubsection{Automatic tuning}

The area of automatic parameter tuning has attracted a lot of attention in
recent years. This is because algorithms have an increasing number of parameters
that are difficult to tune even for experts and because of research into
dynamic algorithm portfolios that benefits from automatic tuning. A survey of
the literature on automatic tuning is outside the scope of this paper, but some
of the approaches that are particularly relevant to this survey are described
below.

Automatic tuning and portfolio selection can be treated separately,
as done in the Hydra portfolio builder \cite{xu_hydra_2010}. Hydra uses
ParamILS \cite{hutter_automatic_2007,hutter_paramils_2009} to automatically
tune algorithms in a SATzilla \cite{xu_satzilla_2008} portfolio. ISAC
\cite{kadioglu_isac_2010} uses GGA \cite{ansotegui_gender-based_2009} to
automatically tune algorithms for clusters of problem instances.

\citeA{minton_automatically_1996} first enumerates all possible rule
applications up to a certain time or size bound. Then, the most promising
configuration is selected using beam search, a form of parallel hill climbing,
that empirically evaluates the performance of each candidate.
\citeA{balasubramaniam_automated_2012} use hill climbing to similarly identify
the most efficient configuration for a constraint solver on a set of problems.
\citeA{terashima-marin_evolution_1999,fukunaga_automated_2002} use genetic
algorithms to evolve promising configurations.

The systems described in the previous paragraph are only of limited suitability
for dynamic algorithm portfolios. They either take a long time to find good
configurations or are restricted in the number or type of parameters.
Interactions between parameters are only taken into account in a limited way.
More recent approaches have focused on overcoming these limitations.

The ParamILS system \cite{hutter_automatic_2007,hutter_paramils_2009} uses
techniques based on local search to identify parameter configurations with good
performance. The authors address over-confidence (overestimating the performance
of a parameter configuration on a test set) and over-tuning (determining a
parameter configuration that is too specific).
\citeA{ansotegui_gender-based_2009} use genetic algorithms to discover
favourable parameter configurations for the algorithms being tuned. The authors
use a racing approach to avoid having to run all generated configurations to
completion. They also note that one of the advantages of the genetic algorithm
approach is that it is inherently parallel.

Both of these approaches are capable of tuning algorithms with a large number of
parameters and possible values as well as taking interactions between parameters
into account. They are used in practice in the Algorithm Selection systems Hydra
and ISAC, respectively. In both cases, they are only used to construct static
portfolios however. More recent approaches focus on exploiting parallelism
\cite<e.g.>{hutter_parallel_2012}.

\medskip

Dynamic portfolios are in general a more fruitful area for Algorithm
Selection research because of the large space of possible decisions. Static
portfolios are usually relatively small and the decision space is amenable for
human exploration. This is not a feasible approach for dynamic portfolios
though. \citeA{minton_automatically_1996} notes that
\begin{quote}
``\textsc{Multi-tac} turned out to have an unexpected advantage in this arena,
due to the complexity of the task. Unlike our human subjects, \textsc{Multi-tac}
experimented with a wide variety of combinations of heuristics. Our human
subjects rarely had the inclination or patience to try many alternatives, and on
at least one occasion incorrectly evaluated alternatives that they did try.''
\end{quote}

\section{Problem solving with portfolios}\label{sec:solving}

Once an algorithm portfolio has been constructed, the way in which it is to be
used has to be decided. There are different considerations to take into account.
The two main issues are as follows.

\begin{description}
\item[What to select]\hfill\\
    Given the full set of algorithms in the portfolio, a subset has to be chosen
    for solving the problem. This subset can consist of only a single algorithm
    that is used to solve the problem to completion, the entire portfolio with
    the individual algorithms interleaved or running in parallel or anything in
    between.
\item[When to select]\hfill\\
    The selection of the subset of algorithms can be made only once before
    solving starts or continuously during search. If the latter is the case,
    selections can be made at well-defined points during search, for example at
    each node of a search tree, or when the system judges it to be necessary to
    make a decision.
\end{description}

Rice's model assumes that only a single algorithm $A \in \mathcal{A}$ is
selected. It implicitly assumes that this selection occurs only once and before
solving the actual problem.

\subsection{What to select}

A common and the simplest approach is to select a single algorithm from the
portfolio and use it to solve the problem completely. This single algorithm has
been determined to be the best for the problem at hand. For example SATzilla
\cite{nudelman_understanding_2004,xu_satzilla-07_2007,xu_satzilla_2008},
\textsc{ArgoSmArT} \cite{nikoli_instance-based_2009}, SALSA
\cite{demmel_self-adapting_2005} and \textsc{Eureka}
\cite{cook_maximizing_1997} do this. The disadvantage of this approach is that
there is no way of mitigating a wrong selection. If an algorithm is chosen that
exhibits bad performance on the problem, the system is ``stuck'' with it and no
adjustments are made, even if all other portfolio algorithms would perform much
better.

An alternative approach is to compute schedules for running (a subset of) the
algorithms in the portfolio. In some approaches, the terms portfolio and
schedule are used synonymously -- all algorithms in the portfolio are selected
and run according to a schedule that allocates time slices to each of them. The
task of Algorithm Selection becomes determining the schedule rather than to
select algorithms.

\citeA{roberts_directing_2006} rank the portfolio algorithms in order of
expected performance and allocate time according to this ranking.
\citeA{howe_exploiting_1999} propose a round-robin schedule that contains all
algorithms in the portfolio. The order of the algorithms is determined by the
expected run time and probability of success. The first algorithm is allocated a
time slice that corresponds to the expected time required to solve the problem.
If it is unable to solve the problem during that time, it and the remaining
algorithms are allocated additional time slices until the problem is solved or a
time limit is reached.

\citeA{pulina_self-adaptive_2009} determine a schedule according to
three strategies. The first strategy is to run all portfolio algorithms for a
short time and if the problem has not been solved after this, run the predicted
best algorithm exclusively for the remaining time. The second strategy runs all
algorithms for the same amount of time, regardless of what the predicted best
algorithm is. The third variation allocates exponentially increasing time slices
to each algorithm such that the total time is again distributed equally among
them. In addition to the three different scheduling strategies, the authors
evaluate four different ways of ordering the portfolio algorithms within a
schedule that range from ranking based on past performance to random. They
conclude that ordering the algorithms based on their past performance and
allocating the same amount of time to all algorithms gives the best overall
performance.

\citeA{omahony_using_2008} optimise the computed schedule with respect to the
probability that the problem will be solved. They use the past performance data
of the portfolio algorithms for this. However, they note that their approach of
using a simple complete search procedure to find this optimal schedule relies on
small portfolio sizes and that ``for a large number of solvers, a more
sophisticated approach would be necessary''.

\citeA{kadioglu_algorithm_2011} formulate the problem of computing a schedule
that solves most problems in a training set in the lowest amount of time as a
resource constrained set covering integer programme. They pursue similar aims as
\citeA{omahony_using_2008} but note that their approach is more efficient and
able to scale to larger schedules. However, their evaluation concludes that the
approach with the best overall performance is to run the predicted best
algorithm for 90\% of the total available time and distribute the remaining 10\%
across the other algorithms in the portfolio according to a static schedule.

\citeA{petrik_statistically_2005} presents a framework for calculating optimal
schedules. The approach is limited by a number of assumptions about the
algorithms and the execution environment, but is applicable to a wide range of
research in the literature. \citeA{petrik_learning_2006,bougeret_combining_2009}
compute an optimal static schedule for allocating fixed time slices to each
algorithm. \citeA{sayag_combining_2006} propose an algorithm to efficiently
compute an optimal schedule for portfolios of fixed size and show that the
problem of generating or even approximating an optimal schedule is
computationally intractable. \citeA{roberts_learned_2007} explore different
strategies for allocating time slices to algorithms. In a serial execution
strategy, each algorithm is run once for an amount of time determined by the
average time to find a solution on previous problems or the time that was
predicted for finding a solution on the current problem. A round-robin strategy
allocates increasing time slices to each algorithm. The length of a time slice
is based on the proportion of successfully solved training problems within this
time. \citeA{gerevini_automatically_2009} compute round-robin schedules
following a similar approach. Not all of their computed schedules contain all
portfolio algorithms. \citeA{streeter_combining_2007} compute a schedule with
the aim of improving the average-case performance. In later work, they compute
theoretical guarantees for the performance of their schedule
\cite{streeter_new_2008}.

\citeA{wu_portfolios_2007} approach scheduling the chosen algorithms in a
different way and assume a fixed limit on the amount of resources an algorithm
can consume while solving a problem. All algorithms are run sequentially for
this fixed amount of time. Similar to \citeA{gerevini_automatically_2009}, they
simulate the performance of different allocations and select the best one based
on the results of these simulations. \cite{fukunaga_genetic_2000} estimates the
performance of candidate allocations through bootstrap sampling.
\citeA{gomes_algorithm_1997,gomes_algorithm_2001} also evaluate the performance
of different candidate portfolios, but take into account how many algorithms can
be run in parallel. They demonstrate that the optimal schedule (in this case the
number of algorithms that are being run) changes as the number of available
processors increases. \citeA{gagliolo_towards_2008} investigate how to allocate
resources to algorithms in the presence of multiple CPUs that allow to run more
than one algorithm in parallel. \citeA{yun_learning_2012} craft portfolios with
the specific aim of running the algorithms in parallel.

\medskip

Related research is concerned with the scheduling of restarts of stochastic
algorithms -- it also investigates the best way of allocating resources. The
paper that introduced algorithm portfolios \cite{huberman_economics_1997} uses a
portfolio of identical stochastic algorithms that are run with different random
seeds. There is a large amount of research on how to determine restart schedules
for randomised algorithms and a survey of this is outside the scope of this
paper. A few approaches that are particularly relevant to Algorithm Selection
and portfolios are mentioned below.

\citeA{horvitz_bayesian_2001} determine the amount of time to allocate to a
stochastic algorithm before restarting it. They use dynamic policies that take
performance predictions into account, showing that it can outperform an optimal
fixed policy.

\citeA{cicirello_max_2005} investigate a restart model model that allocates
resources to an algorithm proportional to the number of times it has been
successful in the past. In particular, they note that the allocated resources
should grow doubly exponentially in the number of successes. Allocation of fewer
resources results in over-exploration (too many different things are tried and
not enough resources given to each) and allocation of more resources in
over-exploitation (something is tried for to too long before moving on to
something different).

\citeA{streeter_restart_2007} compute restart schedules that take the runtime
distribution of the portfolio algorithms into account. They present an approach
that does so statically based on the observed performance on a set of training
problems as well as an approach that learns the runtime distributions as new
problems are solved without a separate training set.

\subsection{When to select}\label{sec:offon}

In addition to whether they choose a single algorithm or compute a schedule,
existing approaches can also be distinguished by whether they operate before the
problem is being solved (offline) or while the problem is being solved (online).
The advantage of the latter is that more fine-grained decisions can be made and
the effect of a bad choice of algorithm is potentially less severe. The price
for this added flexibility is a higher overhead however, as algorithms are
selected more frequently.

Examples of approaches that only make offline decisions include
\citeA{xu_satzilla_2008,minton_automatically_1996,smith_knowledge-based_1992,omahony_using_2008}.
In addition to having no way of mitigating wrong choices, often these will not
even be detected. These approaches do not monitor the execution of the chosen
algorithms to confirm that they conform with the expectations that led to them
being chosen. Purely offline approaches are inherently vulnerable to bad
choices. Their advantage however is that they only need to select an algorithm
once and incur no overhead while the problem is being solved.

Moving towards online systems, the next step is to monitor the execution of an
algorithm or a schedule to be able to intervene if expectations are not met.
\citeA{fink_statistical_1997,fink_how_1998} investigates setting a time
bound for the algorithm that has been selected based on the predicted
performance. If the time bound is exceeded, the solution attempt is abandoned.
More sophisticated systems furthermore adjust their selection if
such a bound is exceeded. \citeA{borrett_adaptive_1996} try to detect behaviour
during search that indicates that the algorithm is performing badly, for example
visiting nodes in a subtree of the search that clearly do not lead to a
solution. If such behaviour is detected, they propose switching the currently
running algorithm according to a fixed replacement list.

\citeA{sakkout_instance_1996} explore the same basic idea. They switch between
two algorithms for solving constraint problems that achieve different levels of
consistency. The level of consistency refers to the amount of search space that
is ruled out by inference before actually searching it. Their approach achieves
the same level of search space reduction as the more expensive algorithm at a
significantly lower cost. This is possible because doing more inference does not
necessarily result in a reduction of the search space in all cases. The authors
exploit this fact by detecting such cases and doing the cheaper inference.
\citeA{stergiou_heuristics_2009} also investigates switching propagation methods
during solving. \citeA{yu_adaptive_2004,yu_adaptive_2006} do not monitor the
execution of the selected algorithm, but instead the values of the features used
to select it. They re-evaluate the selection function when its inputs change.

Further examples of approaches that monitor the execution of the selected
algorithm are \citeA{pulina_self-adaptive_2009,gagliolo_adaptive_2004}, but also
\citeA{horvitz_bayesian_2001} where the offline selection of an algorithm is
combined with the online selection of a restart strategy. An interesting feature
of \citeA{pulina_self-adaptive_2009} is that the authors adapt the model used
for the offline algorithm selection if the actual run time is much higher than
the predicted runtime. In this way, they are not only able to mitigate bad
choices during execution, but also prevent them from happening again.

The approaches that make decisions during search, for example at every node of
the search tree, are necessarily online systems. \citeA{arbelaez_online_2009}
select the best search strategy at checkpoints in the search tree. Similarly,
\citeA{brodley_automatic_1993} recursively partitions the classification problem
to be solved and selects an algorithm for each partition. In this approach, a
lower-level decision can lead to changing the decision at the level above. This
is usually not possible for combinatorial search problems, as decisions at a
higher level cannot be changed easily.

Closely related is the work by
\citeA{lagoudakis_algorithm_2000,lagoudakis_learning_2001}, which partitions the
search space into recursive subtrees and selects the best algorithm from the
portfolio for every subtree. They specifically consider recursive algorithms. At
each recursive call, the Algorithm Selection procedure is invoked. This is a
more natural extension of offline systems than monitoring the execution of the
selected algorithms, as the same mechanisms can be used.
\citeA{samulowitz_learning_2007} also select algorithms for recursively solving
sub-problems.

The PRODIGY system \cite{carbonell_prodigy_1991} selects the next
operator to apply in order to reach the goal state of a planning problem at each
node in the search tree. Similarly, \citeA{langley_learning_1983} learn weights
for operators that can be applied at each search state and select from among
them accordingly.

Most approaches rely on an offline element that makes a decision before search
starts. In the case of recursive calls, this is no different from making a
decision during search however.
\citeA{gagliolo_adaptive_2004,gagliolo_neural_2005,gagliolo_learning_2006} on the
other hand learn the Algorithm Selection model only dynamically while the
problem is being solved. Initially, all algorithms in the portfolio are
allocated the same (small) time slice. As search progresses, the allocation
strategy is updated, giving more resources to algorithms that have exhibited
better performance. The expected fastest algorithm receives half of the total
time, the next best algorithm half of the remaining time and so on.
\citeA{armstrong_dynamic_2006} also rely exclusively on a selection model trained
online in a similar fashion. They evaluate different strategies of allocating
resources to algorithms according to their progress during search. All of these
strategies converge to allocating all resources to the algorithm with the best
observed performance.

\section{Portfolio selectors}\label{sec:selectors}

Research on \emph{how} to select from a portfolio in an Algorithm Selection
system has generated the largest number of different approaches within the
framework of Algorithm Selection. In Rice's framework, it roughly corresponds to
the performance mapping $p(A,x)$, although only few approaches use this exact
formulation. Rice assumes that the performance of a particular algorithm on a
particular problem is of interest. While this is true in general, many
approaches only take this into account implicitly. Selecting the single best
algorithm for a problem for example has no explicit mapping into Rice's
performance measure space $\mathcal{R}^n$ at all. The selection mapping
$S(f(x))$ is also related to the problem of how to select.

There are many different ways a mechanism to select from a portfolio can be
implemented. Apart from accuracy, one of the main requirements for such a
selector is that it is relatively cheap to run -- if selecting an algorithm for
solving a problem is more expensive than solving the problem, there is no point
in doing so. \citeA{vassilevska_confronting_2006} explicitly define the selector
as ``an efficient (polynomial time) procedure''.

There are several challenges associated with making selectors efficient.
Algorithm Selection systems that analyse the problem to be solved, such as
SATzilla, need to take steps to ensure that the analysis does not become too
expensive. Two such measures are the running of a pre-solver and the prediction
of the time required to analyse a problem \cite{xu_satzilla_2008}. The idea
behind the pre-solver is to choose an algorithm with reasonable general
performance from the portfolio and use it to start solving the problem before
starting to analyse it. If the problem happens to be very easy, it will be
solved even before the results of the analysis are available. After a fixed
time, the pre-solver is terminated and the results of the Algorithm Selection
system are used. \citeA{pulina_self-adaptive_2009} use a similar approach and
run all algorithms for a short time in one of their strategies. Only if the
problem has not been solved after that, they move on to the algorithm that was
actually
selected.

Predicting the time required to analyse a problem is a closely related idea. If
the predicted required analysis time is too high, a default algorithm with
reasonable performance is chosen and run on the problem. This technique is
particularly important in cases where the problem is hard to analyse, but easy
to solve. As some systems use information that comes from exploring part of the
search space (cf.\ Section~\ref{sec:features}), this is a very relevant concern
in practice. On some problems, even probing just a tiny part of the search space
may take a very long time.

\citeA{gent_learning_2010,gent_machine_2010} report that using the
misclassification penalty as a weight for the individual problems during
training improves the quality of the predictions. The misclassification penalty
quantifies the ``badness'' of a wrong prediction; in this case as the additional
time required to solve a problem. If an algorithm was chosen that is only
slightly worse than the best one, it has less impact than choosing an algorithm
that is orders of magnitude worse. Using the penalty during training is a way of
guiding the learned model towards the problems where the potential performance
improvement is large.

\medskip

There are many different approaches to how portfolio selectors operate. The
selector is not necessarily an explicit part of the system.
\citeA{minton_automatically_1996} compiles the Algorithm Selection system into a
Lisp programme for solving the original constraint problem. The selection rules
are part of the programme logic.
\citeA{fukunaga_automated_2008,garrido_dvrp_2010} evolve selectors and
combinators of heuristic building blocks using genetic algorithms. The selector
is implicit in the evolved programme.

\subsection{Performance models}

The way the selector operates is closely linked to the way the performance model
of the algorithms in the portfolio is built. In early approaches, the
performance model was usually not learned but given in the form of human expert
knowledge. \citeA{borrett_adaptive_1996,sakkout_instance_1996} use hand-crafted
rules to determine whether to switch the algorithm during solving.
\citeA{allen_selecting_1996} also have hand-crafted rules, but estimate the
runtime performance of an algorithm. More recent approaches sometimes use only
human knowledge as well. \citeA{wei_switching_2008} select a local search
heuristic for solving SAT problems by a hand-crafted rule that considers the
distribution of clause weights. \citeA{tolpin_rational_2011} model the
performance space manually using statistical methods and use this hand-crafted
model to select a heuristic for solving constraint problems.
\citeA{vrakas_learning_2003} learn rules automatically, but then filter them
manually.

A more common approach today is to automatically learn performance models
using Machine Learning on training data. The portfolio algorithms are run on a
set of representative problems and based on these experimental results,
performance models are built. This approach is used by
\citeA{xu_satzilla_2008,pulina_multi-engine_2007,omahony_using_2008,kadioglu_isac_2010,guerri_learning_2004},
to name but a few examples. A drawback of this approach is that the training
time is usually large. \citeA{gagliolo_impact_2006} investigate ways of
mitigating this problem by using censored sampling, which introduces an upper
bound on the runtime of each experiment in the training phase.
\citeA{kotthoff_evaluation_2012} also investigate censored sampling where not
all algorithms are run on all problems in the training phase. Their results show
that censored sampling may not have a significant effect on the performance of
the learned model.

Models can also be built without a separate training phase, but while the
problem is solved. This approach is used by
\citeA{gagliolo_learning_2006,armstrong_dynamic_2006} for example. While this
significantly reduces the time to build a system, it can mean that the result is
less effective and efficient. At the beginning, when no performance models have
been built, the decisions of the selector might be poor. Furthermore, creating
and updating performance models why the problem is being solved incurs an
overhead.

The choice of Machine Learning technique is affected by the way the portfolio
selector operates. Some techniques are more amenable to offline approaches
(e.g.\ linear regression models used by \citeR{xu_satzilla_2008}), while others
lend themselves to online methods (e.g.\ reinforcement learning used by
\citeR{armstrong_dynamic_2006}).

Performance models can be categorised by the type of entity whose performance is
modelled -- the entire portfolio or individual algorithms within it. There are
publications that use both of those categories however
\cite<e.g.>{smith-miles_towards_2008}. In some cases, no performance models as
such are used at all.
\citeA{caseau_meta-heuristic_1999,minton_automatically_1996,balasubramaniam_automated_2012}
run the candidates on a set of test problems and select the one with the best
performance that way for example.
\citeA{gomes_algorithm_1997,wu_portfolios_2007,gerevini_automatically_2009}
simulate the performance of different selections on training data.

\subsubsection{Per-portfolio models}

One automated approach is to learn a performance model of the entire portfolio
based on training data. Usually, the prediction of such a model is the best
algorithm from the portfolio for a particular problem. There is only a weak
notion of an individual algorithm's performance. In Rice's notation for the
performance mapping $P(A,x)$, $A$ is the (subset of the) portfolio instead of an
individual algorithm, i.e.\ $A\subseteq \mathcal{A}$ instead of Rice's $A\in
\mathcal{A}$.

This is used for example by
\citeA{omahony_using_2008,cook_maximizing_1997,pulina_multi-engine_2007,nikoli_instance-based_2009,guerri_learning_2004}.
Again there are different ways of doing this. Lazy approaches do not learn an
explicit model, but use the set of training examples as a case base. For new
problems, the closest problem or the set of $n$ closest problems in the case
base is determined and decisions made accordingly.
\citeA{wilson_case-based_2000,pulina_multi-engine_2007,omahony_using_2008,nikoli_instance-based_2009,gebruers_making_2004,malitsky_non-model-based_2011}
use nearest-neighbour classifiers to achieve this. Apart from the conceptual
simplicity, such an approach is attractive because it does not try to abstract
from the examples in the training data. The problems that Algorithm Selection
techniques are applied to are usually complex and factors that affect the
performance are hard to understand. This makes it hard to assess whether a
learned abstract model is appropriate and what its requirements and limitations
are.

Explicitly-learned models try to identify the concepts that affect performance
for a given problem. This acquired knowledge can be made explicit to improve the
understanding of the researchers of the problem domain. There are several
Machine Learning techniques that facilitate this, as the learned models are
represented in a form that is easy to understand by humans.
\citeA{carbonell_prodigy_1991,gratch_composer_1992,brodley_automatic_1993,vrakas_learning_2003}
learn classification rules that guide the selector. \citeA{vrakas_learning_2003}
note that the decision to use a classification rule leaner was not so much
guided by the performance of the approach, but the easy interpretability of the
result.
\citeA{langley_learning_1983,epstein_adaptive_2002,nareyek_choosing_2001} learn
weights for decision rules to guide the selector towards the best algorithms.
\citeA{cook_maximizing_1997,guerri_learning_2004,guo_learning-based_2004,roberts_directing_2006,bhowmick_application_2006,gent_learning_2010}
go one step further and learn decision trees. \citeA{guo_learning-based_2004}
again note that the reason for choosing decision trees was not primarily the
performance, but the understandability of the result.
\citeA{pfahringer_meta-learning_2000} show the set of learned rules in the paper
to illustrate its compactness. Similarly, \citeA{gent_learning_2010} show their
final decision tree in the paper.

Some approaches learn probabilistic models that take uncertainty and variability
into account. \citeA{gratch_composer_1992} use a probabilistic model to learn
control rules. The probabilities for candidate rules being beneficial are
evaluated and updated on a training set until a threshold is reached. This
methodology is used to avoid having to evaluate candidate rules on larger
training sets, which would show their utility more clearly but be more
expensive. \citeA{demmel_self-adapting_2005} learn multivariate Bayesian
decision rules. \citeA{carchrae_low-knowledge_2004} learn a Bayesian classifier
to predict the best algorithm after a certain amount of time.
\citeA{stern_collaborative_2010} learn Bayesian models that incorporate
collaborative filtering. \citeA{domshlak_max_2010} learn decision rules using
na\"ive Bayes classifiers.
\citeA{lagoudakis_algorithm_2000,petrik_statistically_2005} learn performance
models based on Markov Decision Processes. \citeA{kotthoff_evaluation_2012} use
statistical relational learning to predict the ranking of the algorithms in the
portfolio on a particular problem. None of these approaches make explicit use of
the uncertainty attached to a decision though.

Other approaches include support vector machines
\cite{hough_modern_2006,arbelaez_online_2009}, reinforcement learning
\cite{armstrong_dynamic_2006}, neural networks \cite{gagliolo_neural_2005},
decision tree ensembles \cite{hough_modern_2006}, ensembles of general
classification algorithms \cite{kotthoff_ensemble_2010}, boosting
\cite{bhowmick_application_2006}, hybrid approaches that combine regression and
classification \cite{kotthoff_hybrid_2012}, multinomial logistic regression
\cite{samulowitz_learning_2007}, self-organising maps
\cite{smith-miles_towards_2008} and clustering
\cite{stamatatos_learning_2009,stergiou_heuristics_2009,kadioglu_isac_2010}.
\citeA{sayag_combining_2006,streeter_combining_2007} compute schedules for
running the algorithms in the portfolio based on a statistical model of the
problem instance distribution and performance data for the algorithms. This is
not an exhaustive list, but focuses on the most prominent approaches and
publications. Within a single family of approaches, such as decision trees,
there are further distinctions that are outside the scope of this paper, such as
the type of decision tree inducer.

\citeA{arbelaez_online_2009} discuss a technical issue related to the
construction of per-portfolio performance models. A particular algorithm often
exhibits much better performance in general than other algorithms on a
particular instance distribution. Therefore, the training data used to learn
the performance model will be skewed towards that algorithm. This can be a
problem for Machine Learning, as always predicting this best algorithm might
have a very high accuracy already, making it very hard to improve on. The
authors mention two means of mitigating this problem. The training set can be
\emph{under-sampled}, where examples where the best overall algorithm performs
best are deliberately omitted. Alternatively, the set can be \emph{over-sampled}
by artificially increasing the number of examples where another algorithm is
better.

\subsubsection{Per-algorithm models}

A different approach is to learn performance models for the individual
algorithms in the portfolio. The predicted performance of an algorithm on a
problem can be compared to the predicted performance of the other portfolio
algorithms and the selector can proceed based on this. The advantage of this
approach is that it is easier to add and remove algorithms from the portfolio --
instead of having to retrain the model for the entire portfolio, it suffices to
train a model for the new algorithm or remove one of the trained models.
Most approaches only rely on the order of predictions being correct. It does not
matter if the prediction of the performance itself is wildly inaccurate as long
as it is correct relative to the other predictions.

This is the approach that is implicitly assumed in Rice's framework. The
prediction is the performance mapping $P(A,x)$ for an algorithm
$A\in\mathcal{A}$ on a problem $x\in\mathcal{P}$. Models for each algorithm in
the portfolio are used for example by
\citeA{xu_satzilla_2008,howe_exploiting_1999,allen_selecting_1996,lobjois_branch_1998,gagliolo_learning_2006}.

A common way of doing this is to use regression to directly predict the
performance of each algorithm. This is used by
\citeA{xu_satzilla_2008,howe_exploiting_1999,leyton-brown_learning_2002,haim_restart_2009,roberts_learned_2007}.
The performance of the algorithms in the portfolio is evaluated on a set of
training problems, and a relationship between the characteristics of a problem
and the performance of an algorithm derived. This relationship usually has the
form of a simple formula that is cheap to compute at runtime.

\citeA{silverthorn_latent_2010} on the other hand learn latent class models of
unobserved variables to capture relationships between solvers, problems and run
durations. Based on the predictions, the expected utility is computed and used
to select an algorithm. \citeA{sillito_improvements_2000} surveys sampling
methods to estimate the cost of solving constraint problems.
\citeA{watson_empirical_2003} models the behaviour of local search algorithms
with Markov chains.

Another approach is to build statistical models of an algorithm's performance
based on past observations. \citeA{weerawarana_pythia_1996} use Bayesian belief
propagation to predict the runtime of a particular algorithm on a particular
problem. Bayesian inference is used to determine the class of a problem and the
closest case in the knowledge base. A performance profile is extracted from that
and used to estimate the runtime. The authors also propose an alternative
approach that uses neural nets. \citeA{fink_statistical_1997,fink_how_1998}
computes the expected gain for time bounds based on past success times. The
computed values are used to choose the algorithm and the time bound for running
it. \citeA{brazdil_comparison_2000} compare algorithm rankings based on different
past performance statistics. Similarly, \citeA{leite_using_2010} maintain a
ranking based on past performance. \citeA{cicirello_max_2005} propose a bandit
problem model that governs the allocation of resources to each algorithm in the
portfolio. \citeA{wang_optimizing_2007} also use a bandit model, but furthermore
evaluate a Q-learning approach, where in addition to bandit model rewards, the
states of the system are taken into account.
\citeA{gomes_algorithm_1997,wu_portfolios_2007,gerevini_automatically_2009} use
the past performance of algorithms to simulate the performance of different
algorithm schedules and use statistical tests to select one of the schedules.

\subsubsection{Hierarchical models}\label{hierarchical}

There are some approaches that combine several models into a hierarchical
performance model. There are two basic types of hierarchical models. One type
predicts additional \emph{properties of the problem} that cannot be measured
directly or are not available without solving the problem. The other type makes
\emph{intermediate predictions} that do not inform Algorithm Selection directly,
but rather the final predictions.

\citeA{xu_hierarchical_2007} use sparse multinomial logistic regression to
predict whether a SAT problem instance is satisfiable and, based on that
prediction, use a logistic regression model to predict the runtime of each
algorithm in the portfolio. \citeA{haim_restart_2009} also predict the
satisfiability of a SAT instance and then choose an algorithm from a portfolio.
Both report that being able to distinguish between satisfiable and unsatisfiable
problems enables performance improvements. The satisfiability of a problem is a
property that needs to be \emph{predicted} in order to be useful for Algorithm
Selection. If the property is \emph{computed} (i.e.\ the problem is solved),
there is no need to perform Algorithm Selection anymore.

\citeA{gent_machine_2010} use classifiers to first decide on the level of
consistency a constraint propagator should achieve and then on the actual
implementation of the propagator that achieves the selected level of
consistency. A different publication that uses the same data set does not make
this distinction however \cite{kotthoff_ensemble_2010}, suggesting that the
performance benefits are not significant in practice.

Such hierarchical models are only applicable in a limited number of scenarios,
which explains the comparatively small amount of research into them. For many
application domains, only a single property needs to be predicted and can be
predicted without intermediate steps with sufficient accuracy.
\citeA{kotthoff_hybrid_2012} proposes a hierarchical approach that is
domain-independent. He uses the performance predictions of regression models as
input to a classifier that decides which algorithm to choose and demonstrates
performance improvements compared to selecting an algorithm directly based on
the predicted performance. The idea is very similar to that of \emph{stacking}
in Machine Learning \citeA{wolpert_stacked_1992}.

\subsubsection{Selection of model learner}

Apart from the different types of performance models, there are different
Machine Learning algorithms that can be used to learn a particular kind of
model. While most of the approaches mentioned here rely on a single way of doing
this, some of the research compares different methods.

\citeA{xu_satzilla_2008} mention that, in addition to the chosen ridge regression
for predicting the runtime, they explored using lasso regression, support vector
machines and Gaussian processes. They chose ridge regression not because it
provided the most accurate predictions, but the best trade-off between accuracy
and cost to make the prediction. \citeA{weerawarana_pythia_1996} propose an
approach that uses neural networks in addition to the Bayesian belief
propagation approach they describe initially. \citeA{cook_maximizing_1997}
compare different decision tree learners, a Bayesian classifier, a nearest
neighbour approach and a neural network. They chose the C4.5 decision tree
inducer because even though it may be outperformed by a neural network, the
learned trees are easily understandable by humans and may provide insight into
the problem domain. \citeA{leyton-brown_learning_2002} compare several versions
of linear and non-linear regression. \citeA{hutter_performance_2006} report
having explored support vector machine regression, multivariate adaptive
regression splines (MARS) and lasso regression before deciding to use the linear
regression approach of \citeA{leyton-brown_learning_2002}. They also report
experimental results with sequential Bayesian linear regression and Gaussian
Process regression. \citeA{guo_algorithm_2003,guo_learning-based_2004} explore
using decision trees, na\"ive Bayes rules, Bayesian networks and meta-learning
techniques. They also chose the C4.5 decision tree inducer because it is one of
the top performers and creates models that are easy to understand and quick to
execute. \citeA{gebruers_using_2005} compare nearest neighbour classifiers,
decision trees and statistical models. They show that a nearest neighbour
classifier outperforms all the other approaches on their data sets.

\citeA{hough_modern_2006} use decision tree ensembles and support vector
machines. \citeA{bhowmick_application_2006} investigate alternating decision
trees and various forms of boosting, while \citeA{pulina_multi-engine_2007} use
decision trees, decision rules, logistic regression and nearest neighbour
approaches. They do not explicitly choose one of these methods in the paper, but
their Algorithm Selection system AQME uses a nearest neighbour classifier by
default. \citeA{roberts_learned_2007} use 32 different Machine Learning
algorithms to predict the runtime of algorithms and probability of success. They
attempt to provide explanations for the performance of the methods they have
chosen in \citeA{roberts_what_2008}. \citeA{silverthorn_latent_2010} compare the
performance of different latent class models. \citeA{gent_machine_2010} evaluate
the performance of 19 different Machine Learning classifiers on an Algorithm
Selection problem in constraint programming. The investigation is extended to
include more Machine Learning algorithms as well as different performance models
and more problem domains in \citeA{kotthoff_evaluation_2012}. They identify
several Machine Learning algorithms that show particularly good performance
across different problem domains, namely linear regression and alternating
decision trees. They do not consider issues such as how easy the models are to
understand or how efficient they are to compute.

Only
\citeA{guo_learning-based_2004,gebruers_using_2005,hough_modern_2006,pulina_multi-engine_2007,silverthorn_latent_2010,gent_machine_2010,kotthoff_evaluation_2012}
quantify the differences in performance of the methods they used. The other
comparisons give only qualitative evidence. Not all comparisons choose one of
the approaches over the other or provide sufficient detail to enable the reader
to do so. In cases where a particular technique is chosen, performance is often
not the only selection criterion. In particular, the ability to understand a
learned model plays a significant role.

\subsection{Types of predictions}

The way of creating the performance model of a portfolio or its algorithms is
not the only choice researchers face. In addition, there are different
predictions the performance model can make to inform the decision of the
selector of a subset of the portfolio algorithms. The type of decision is
closely related to the learned performance model however. The prediction can be
a single categorical value -- the algorithm to choose. This type of prediction
is usually the output of per-portfolio models and used for example in
\citeA{gent_learning_2010,cook_maximizing_1997,pulina_multi-engine_2007,nikoli_instance-based_2009,guerri_learning_2004}.
The advantage of this simple prediction is that it determines the choice of
algorithm without the need to compare different predictions or derive further
quantities. One of its biggest disadvantages however is that there is no
flexibility in the way the system runs or even the ability to monitor the
execution for unexpected behaviour.

A different approach is to predict the runtime of the individual algorithms in
the portfolio. This requires per-algorithm models. For example
\citeA{horvitz_bayesian_2001,petrik_statistically_2005,silverthorn_latent_2010}
do this. \citeA{xu_satzilla_2008} do not predict the runtime itself, but the
logarithm of the runtime. They note that,
\begin{quote}
``In our experience, we have found this log transformation of runtime to be very
important due to the large variation in runtimes for hard combinatorial
problems.''
\end{quote}
\citeA{kotthoff_evaluation_2012} also compare predicting the runtime itself and
the log thereof, but find no significant difference between the two.
\citeA{kotthoff_hybrid_2012} however also reports better results with the
logarithm.

\citeA{allen_selecting_1996} estimate the runtime by proxy by predicting the
number of constraint checks. \citeA{lobjois_branch_1998} estimate the runtime by
predicting the number of search nodes to explore and the time per node.
\citeA{lagoudakis_algorithm_2000} talk of the \emph{cost} of selecting a
particular algorithm, which is equal to the time it takes to solve the problem.
\citeA{nareyek_choosing_2001} uses the \emph{utility} of a choice to make his
decision. The utility is an abstract measure of the ``goodness'' of an algorithm
that is adapted dynamically. \citeA{tolpin_rational_2011} use the \emph{value of
information} of selecting an algorithm, defined as the amount of time saved by
making this choice. \citeA{xu_satzilla2009_2009} predict the \emph{penalized
average runtime score}, a measure that combines runtime with possible timeouts.
This approach aims to provide more realistic performance predictions when
runtimes are capped.

More complex predictions can be made, too. In most cases, these are made by
combining simple predictions such as the runtime performance.
\citeA{brazdil_comparison_2000,soares_meta-learning_2004,leite_using_2010}
produce rankings of the portfolio algorithms. \citeA{kotthoff_evaluation_2012}
use statistical relational learning to directly predict the ranking instead of
deriving it from other predictions.
\citeA{howe_exploiting_1999,gagliolo_adaptive_2004,gagliolo_learning_2006,roberts_directing_2006,omahony_using_2008}
predict resource allocations for the algorithms in the portfolios.
\citeA{gebruers_using_2005,little_capturing_2002,borrett_context_2001} consider
selecting the most appropriate formulation of a constraint problem.
\citeA{smith_knowledge-based_1992,brewer_high-level_1995,wilson_case-based_2000,balasubramaniam_automated_2012}
select algorithms and data structures to be used in a software system.

Some types of predictions require online approaches that make decisions during
search.
\citeA{borrett_adaptive_1996,sakkout_instance_1996,carchrae_low-knowledge_2004,armstrong_dynamic_2006}
predict when to switch the algorithm used to solve a problem.
\citeA{horvitz_bayesian_2001} predict whether to restart an algorithm.
\citeA{lagoudakis_algorithm_2000,lagoudakis_learning_2001} predict the cost to
solve a sub-problem. However, most online approaches make predictions that can
also be used in offline settings, such as the best algorithm to proceed with.

The primary selection criteria and prediction for
\citeA{soares_meta-learning_2004} and \citeA{leite_using_2010} is the quality of
the solution an algorithm produces rather than the time it takes the algorithm
to find that solution. In addition to the primary selection criteria, a number
of approaches predict secondary criteria.
\citeA{howe_exploiting_1999,fink_how_1998,roberts_learned_2007} predict the
probability of success for each algorithm. \citeA{weerawarana_pythia_1996}
predict the quality of a solution.

In Rice's model, the prediction of an Algorithm Selection system is the
performance $p\in\mathcal{R}^n$ of an algorithm. This abstract notion does not
rely on time and is applicable to many approaches. It does not fit techniques
that predict the portfolio algorithm to choose or more complex measures such as
a schedule however. As Rice developed his approach long before the advent of
algorithm portfolios, it should not be surprising that the notion of the
performance of individual algorithms as opposed to sets of algorithms dominates.
The model is sufficiently general to be able to accommodate algorithm portfolios
with only minor modifications to the overall framework however.

\section{Features}\label{sec:features}

The different types of performance models described in the previous sections
usually use features to inform their predictions. Features are an integral part
of systems that do Machine Learning. They characterise the inputs, such as the
problem to be solved or the algorithm employed to solve it, and facilitate
learning the relationship between the inputs and the outputs, such as the time
it will take the algorithm to solve the problem. In Rice's model, features
$f(x)$ for a particular problem $x$ are extracted from the feature space
$\mathcal{F}$.

The selection of the most suitable features is an important part of the design
of Algorithm Selection systems. There are different types of features
researchers can use and different ways of computing these. They can be
categorised according to two main criteria.

First, they can be categorised according to how much background knowledge a
researcher needs to have to be able to use them. Features that require no or
very little knowledge of the application domain are usually very general and can
be applied to new Algorithm Selection problems with little or no modification.
Features that are specific to a domain on the other hand may require the
researcher building the Algorithm Selection system to have a thorough
understanding of the domain. These features usually cannot be applied to other
domains, as they may be non-existent or uninformative in different contexts.

The second way of distinguishing different classes of features is according to
when and how they are computed. Features can be computed \emph{statically},
i.e.\ before the search process starts, or \emph{dynamically}, i.e.\ during
search. These two categories roughly align with the offline and online
approaches to portfolio problem solving described in Section~\ref{sec:solving}.

\citeA{smith-miles_measuring_2012} present a survey that focuses on what
features can be used for Algorithm Selection. This paper categorises the
features used in the literature.

\subsection{Low and high-knowledge features}

In some cases, researchers use a large number of features that are specific to
the particular problem domain they are interested in, but there are also
publications that only use a single, general feature -- the performance of a
particular algorithm on past problems.
\citeA{gagliolo_adaptive_2004,petrik_statistically_2005,cicirello_max_2005,streeter_combining_2007,silverthorn_latent_2010},
to name but a few examples, use this approach to build statistical
performance models of the algorithms in their portfolios. The underlying
assumption is that all problems are similar with respect to the relative
performance of the algorithms in the portfolio -- the algorithm that has done
best in the past has the highest chance of performing best in the future.

Approaches that build runtime distribution models for the portfolio algorithms
usually do not select a single algorithm for solving a problem, but rather use
the distributions to compute resource allocations for the individual portfolio
algorithms. The time allocated to each algorithm is proportional to its past
performance.

Other sources of features that are not specific to a particular problem domain
are more fine-grained measures of past performance or measures that characterise
the behaviour of an algorithm during search. \citeA{langley_learningd_1983} for
example determines whether a search step performed by a particular algorithm is
good, i.e.\ leading towards a solution, or bad, i.e.\ straying from the path to
a solution if the solution is known or revisiting an earlier search state if the
solution is not known. \citeA{gomes_algorithm_1997,gomes_algorithm_2001} use the
runtime distributions of algorithms over the size of a problem, as measured by
the number of backtracks. \citeA{fink_how_1998} uses the past success times of
an algorithm as candidate time bounds on new problems.
\citeA{brazdil_comparison_2000} do not consider the runtime, but the error rate
of algorithms. \citeA{gerevini_automatically_2009} use both computation time and
solution quality.

\citeA{beck_simple_2004,carchrae_low-knowledge_2004,carchrae_applying_2005}
evaluate the performance also during search. They explicitly focus on features
that do not require a lot of domain knowledge. \citeA{beck_simple_2004} note
that,
\begin{quote}
``While existing algorithm selection techniques have shown impressive results,
their knowl\-edge-intensive nature means that domain and algorithm expertise is
necessary to develop the models. The overall requirement for expertise has not
been reduced: it has been shifted from algorithm selection to predictive model
building.''
\end{quote}
They do, like several other approaches, assume \emph{anytime} algorithms --
after search has started, the algorithm is able to return the best solution
found so far at any time. The features are based on how search progresses and
how the quality of solutions is improved by algorithms. While this does not
require any knowledge about the application domain, it is not applicable in
cases when only a single solution is sought.

Most approaches learn models for the performance on particular problems and do
not use past performance as a feature, but to inform the prediction to be made.
Considering problem features facilitates a much more nuanced approach than a
broad-brush general performance model. This is the classic supervised Machine
Learning approach -- given the correct prediction derived from the behaviour on
a set of training problems, learn a model that enables to make this prediction.

The features that are considered to learn the model are specific to the problem
domain or even a subset of the problem domain to varying extents. For
combinatorial search problems, the most commonly used basic features include,
\begin{itemize}
\item the number of variables,
\item properties of the variable domains, i.e.\ the list of possible
    assignments,
\item the number of clauses in SAT, the number of constraints in constraint
    problems, the number of goals in planning,
\item the number of clauses/constraints/goals of a particular type (for example
    the number of \texttt{alldifferent} constraints, \citeR{gent_machine_2010}),
\item ratios of several of the above features and summary statistics.
\end{itemize}
Such features are used for example in
\citeA{omahony_using_2008,pulina_multi-engine_2007,weerawarana_pythia_1996,howe_exploiting_1999,xu_satzilla_2008}.

Other sources of features include
the generator that produced the problem to be
solved \cite{horvitz_bayesian_2001}, the runtime environment
\cite{armstrong_dynamic_2006}, structures derived from the problem such as the
primal graph of a constraint problem
\cite{gebruers_making_2004,guerri_learning_2004,gent_learning_2010}, specific
parts of the problem model such as variables \cite{epstein_collaborative_2001},
the algorithms in the portfolio themselves \cite{hough_modern_2006} or
the domain of the problem to be solved \cite{carbonell_prodigy_1991},
\citeA{gerevini_automatically_2009} rely on the problem domain as the only
problem-specific feature and select based on past performance data for the
particular domain. \citeA{beck_dynamic_2000} consider not only the values of
properties of a problem, but the changes of those values while the problem is
being solved. \citeA{smith_knowledge-based_1992} consider features of abstract
representations of the algorithms. \citeA{yu_adaptive_2004,yu_adaptive_2006} use
features that represent technical details of the behaviour of an algorithm on a
problem, such as the type of computations done in a loop.

Most approaches use features that are applicable to all problems of the
application domain they are considering. However, \citeA{horvitz_bayesian_2001}
use features that are not only specific to their application domain, but also to
the specific family of problems they are tackling, such as the variance of
properties of variables in different columns of Latin squares. They note that,
\begin{quote}
``\ldots{}the inclusion of such domain-specific features was important in
learning strongly predictive models.''
\end{quote}

\subsection{Static and dynamic features}

In most cases, the approaches that use a large number of domain-specific
features compute them \emph{offline}, i.e.\ before the solution process starts
(cf.\ Section~\ref{sec:offon}). Examples of publications that only use such
static features are
\citeA{leyton-brown_learning_2002,pulina_multi-engine_2007,guerri_learning_2004}.

An implication of using static features is that the decisions of the Algorithm
Selection system are only informed by the performance of the algorithms on past
problems. Only dynamic features allow to take the performance on the current
problem into account. This has the advantage that remedial actions can be taken
if the problem is unlike anything seen previously or the predictions are wildly
inaccurate for another reason.

A more flexible approach than to rely purely on static features is to
incorporate features that can be determined statically, but try to estimate the
performance on the current problem. Such features are computed by probing the
search space. This approach relies on the performance probes being sufficiently
representative of the entire problem and sufficiently equal across the different
evaluated algorithms. If an algorithm is evaluated on a part of the search space
that is much easier or harder than the rest, a misleading impression of its true
performance may result.

Examples of systems that combine static features of the problem to be solved
with features derived from probing the search space are
\citeA{xu_satzilla_2008,gent_learning_2010,omahony_using_2008}. There are also
approaches that use only probing features. We term this \emph{semi-static}
feature computation because it happens before the actual solving of the problem
starts, but parts of the search space are explored during feature extraction.
Examples include
\citeA{allen_selecting_1996,beck_simple_2004,lobjois_branch_1998}.

The idea of probing the search space is related to \emph{landmarking}
\cite{pfahringer_meta-learning_2000}, where the performance of a set of initial
algorithms (the \emph{landmarkers}) is linked to the performance of the set of
algorithms to select from. The main consideration when using this technique is
to select landmarkers that are computationally cheap. Therefore, they are
usually versions of the portfolio algorithms that have either been simplified or
are run only on a subset of the data the selected algorithm will run on.

While the work done during probing explores part of the search space and could
be used to speed search up subsequently by avoiding to revisit known areas,
almost no research has been done into this. \citeA{beck_simple_2004} run all
algorithms in their (small) portfolio on a problem for a fixed time and select
the one that has made the best progress. The chosen algorithm resumes its
earlier work, but no attempt is made to avoid duplicating work done by the
other algorithms. To the best of our knowledge, there exist no systems that
attempt to avoid redoing work performed by a different algorithm during the
probing stage.

For successful systems, the main source of performance improvements is the
selection of the right algorithm using the features computed through probing. As
the time to compute the features is usually small compared to the runtime
improvements achieved by Algorithm Selection, using the results of probing
during search to avoid duplicating work does not have the potential to achieve
large additional performance improvements.

The third way of computing features is to do so \emph{online}, i.e.\ while
search is taking place. These dynamic features are computed by an execution
monitor that adapts or changes the algorithm during search based on its
performance. Approaches that rely purely on dynamic features are for example
\citeA{borrett_adaptive_1996,nareyek_choosing_2001,stergiou_heuristics_2009}.

There are many different features that can be computed during search.
\citeA{minton_automatically_1996} determines how closely a generated heuristic
approximates a generic target heuristic by checking the heuristic choices at
random points during search. He selects the one with the closest match.
Similarly, \citeA{nareyek_choosing_2001} learn how to select heuristics during
the search process based on their performance. \citeA{armstrong_dynamic_2006} use
an agent-based model that rewards good actions and punishes bad actions based on
computation time. \citeA{kuefler_using_2008} follow a very similar
approach that also takes success or failure into account.

\citeA{carchrae_low-knowledge_2004,carchrae_applying_2005} monitor the solution
quality during search. They decide whether to switch the current algorithm based
on this by changing the allocation of resources. \citeA{wei_switching_2008}
monitor a feature that is specific to their application domain, the distribution
of clause weights in SAT, during search and use it to decide whether to switch a
heuristic. \citeA{stergiou_heuristics_2009} monitors propagation events in a
constraint solver to a similar aim. \citeA{caseau_meta-heuristic_1999} evaluate
the performance of candidate algorithms in terms of number of calls to a
specific high-level procedure. They note that in contrast to using the runtime,
their approach is machine-independent.

\subsection{Feature selection}

The features used for learning the Algorithm Selection model are crucial to its
success. Uninformative features might prevent the model learner from recognising
the real relation between problem and performance or the most important
feature might be missing. Many researchers have recognised this problem.

\citeA{howe_exploiting_1999} manually select the most important features. They
furthermore take the unique approach of learning one model per feature for
predicting the probability of success and combine the predictions of the models.
\citeA{leyton-brown_learning_2002,xu_satzilla_2008} perform automatic feature
selection by greedily adding features to an initially empty set. In addition to
the basic features, they also use the pairwise products of the features.
\citeA{pulina_multi-engine_2007} also perform automatic greedy feature
selection, but do not add the pairwise products.
\citeA{kotthoff_evaluation_2012} automatically select the most important subset
of the original set of features, but conclude that in practice the performance
improvement compared to using all features is not significant.
\citeA{wilson_case-based_2000} use genetic algorithms to determine the
importance of the individual features. \citeA{petrovic_case-based_2002} evaluate
subsets of the features they use and learn weights for each of them.
\citeA{roberts_what_2008} consider using a single feature and automatic
selection of a subset of all features. \citeA{guo_learning-based_2004} and
\citeA{kroer_feature_2011} also use techniques for automatically determining the
most predictive subset of features. \citeA{kotthoff_hybrid_2012} compares the
performance of ten different sets of features.

It is not only important to use informative features, but also features that are
cheap to compute. If the cost of computing the features and making the decision
is too high, the performance improvement from selecting the best algorithm might
be eroded. \citeA{xu_satzilla2009_2009} predict the feature computation time for
a given problem and fall back to a default selection if it is too high to avoid
this problem. They also limit the computation time for the most expensive
features as well as the total time allowed to compute features.
\citeA{bhowmick_towards_2009} consider the computational complexity of
calculating problem features when selecting the features to use. They show that
while achieving comparable accuracy to the full set of features, the subset of
features selected by their method is significantly cheaper to compute.
\citeA{gent_learning_2010} explicitly exclude features that are expensive to
compute.

\section{Application domains}\label{sec:domains}

The approaches for solving the Algorithm Selection Problem that have been
surveyed here are usually not specific to a particular application domain,
within combinatorial search problems or otherwise. Nevertheless this survey
would not be complete without a brief exposition of the various contexts in
which Algorithm Selection techniques have been applied.

Over the years, Algorithm Selection systems have been used in many different
application domains. These range from Mathematics, e.g.\ differential equations
\cite{kamel_odexpert_1993,weerawarana_pythia_1996}, linear algebra
\cite{demmel_self-adapting_2005} and linear systems
\cite{bhowmick_application_2006,kuefler_using_2008}, to the selection of
algorithms and data structures in software design
\cite{smith_knowledge-based_1992,cahill_knowledge-based_1994,brewer_high-level_1995,wilson_case-based_2000}.
A very common application domain are combinatorial search problems such as SAT
\cite{xu_satzilla_2008,lagoudakis_learning_2001,silverthorn_latent_2010},
constraints
\cite{minton_automatically_1996,epstein_adaptive_2002,omahony_using_2008},
Mixed Integer Programming \cite{xu_hydra-mip_2011},
Quantified Boolean Formulae
\cite{pulina_self-adaptive_2009,stern_collaborative_2010}, planning
\cite{carbonell_prodigy_1991,howe_exploiting_1999,vrakas_learning_2003},
scheduling \cite{beck_dynamic_2000,beck_simple_2004,cicirello_max_2005},
combinatorial auctions
\cite{leyton-brown_learning_2002,gebruers_making_2004,gagliolo_learning_2006},
Answer Set Programming \cite{gebser_portfolio_2011},
the Travelling Salesperson Problem \cite{fukunaga_genetic_2000}
and general search algorithms
\cite{langley_learningd_1983,cook_maximizing_1997,lobjois_branch_1998}.

Other domains include Machine Learning
\cite{soares_meta-learning_2004,leite_using_2010}, the most probable
explanation problem \cite{guo_learning-based_2004}, parallel reduction
algorithms \citeA{yu_adaptive_2004,yu_adaptive_2006} and simulation
\cite{wang_optimizing_2007,ewald_selecting_2010}. It should be noted that a
significant part of Machine Learning research is concerned with developing
Algorithm Selection techniques; the publications listed in this paragraph are
the most relevant that use the specific techniques and framework surveyed here.

Some publications consider more than one application domain.
\citeA{stern_collaborative_2010} choose the best algorithm for Quantified
Boolean Formulae and combinatorial auctions.
\citeA{allen_selecting_1996,kroer_feature_2011} look at SAT and constraints.
\citeA{gomes_algorithm_2001} consider SAT and Mixed Integer Programming. In
addition to these two domains, \citeA{kadioglu_isac_2010} also investigate set
covering problems. \citeA{streeter_new_2008} apply their approach to SAT,
Integer Programming and planning.
\citeA{gagliolo_algorithm_2011,kotthoff_evaluation_2012,kotthoff_hybrid_2012}
compare the performance across Algorithm Selection problems from constraints,
Quantified Boolean Formulae and SAT.

In most cases, researchers take some steps to adapt their approaches to the
application domain. This is usually done by using domain-specific features, such
as the number of constraints and variables in constraint programming. In
principle, this is not a limitation of the proposed techniques as those features
can be exchanged for ones that are applicable in other application domains.
While the overall approach remains valid, the question of whether the
performance would be acceptable arises. \citeA{kotthoff_evaluation_2012}
investigate how specific techniques perform across several domains with the aim
of selecting the one with the best overall performance. There are approaches
that have been tailored to a specific application domain to such an extent that
the technique cannot be used for other applications. This is the case for
example in the case of hierarchical models for SAT
\cite{xu_hierarchical_2007,haim_restart_2009}.

\section{Current and future directions}\label{sec:directions}

Research into the Algorithm Selection Problem is ongoing. Many aspects of
Algorithm Selection in various contexts have been explored already. Current
research is extending and refining existing approaches, as well as exploring new
directions. Some of them are listed below, in no particular order.

\subsection{Use of more sophisticated Machine Learning techniques}

Most of the research to date has focused on predicting either the best algorithm
in a portfolio or the performance of an algorithm on a particular problem. In
some cases, these simple predictions are used to generate more complex outputs,
such as a schedule according to which to run the algorithms.
\citeA{kotthoff_evaluation_2012} have started exploring Machine Learning
techniques to predict such complex outputs more directly, but their results are
not competitive with other approaches.

A related direction is to explore the use of generic Machine Learning techniques
that can be applied to many approaches to improve performance.
\citeA{kotthoff_hybrid_2012} for example explores this.
\citeA{xu_evaluating_2012} analyse the performance of a portfolio and the
contributions of its constituent algorithms. The results of such an analysis
could be used to inform the choice of suitable Machine Learning techniques.
\citeA{smith-miles_measuring_2012} focus on identifying features that are
suitable for Machine Learning in Algorithm Selection.

This raises the question of what type of Machine Learning to use in general.
While this has long been a research topic in Machine Learning research, there is
almost no research that applies such knowledge to Algorithm Selection. This
problem is in particular interesting as the authors of the SATzilla system
decided to fundamentally change the type of Machine Learning they use in a
recent publication \cite{xu_hydra-mip_2011}.

\subsection{Exploitation of parallelism}

Many researchers acknowledge at least implicitly that their approaches can be
parallelised across the many cores that modern computers provide. Current
research has started to focus on explicitly exploiting parallelism
\cite<e.g.>{gagliolo_towards_2008,yun_learning_2012,hutter_parallel_2012}. Apart
from technical considerations, one of the main issues is that the composition of
a good algorithm portfolio changes with the number of processors available to
run those algorithms.

There remain challenges that have been largely ignored so far however. As an
example, some portfolio algorithms may be able to take advantage of specialised
processing units such as GPUs while others are not. This would place
restrictions on how the algorithms can be run in parallel. Given the current
trend to have more powerful GPUs with increasing numbers of processing elements
in off-the-shelf computers, we expect this direction of research to become more
prominent.

\subsection{Application to new domains}

Even though Algorithm Selection techniques have been applied to many domains,
especially in Artificial Intelligence, there remain many more that might benefit
from its research. Recently, Algorithm Selection techniques have been applied to
Answer Set Programming for example \cite{gebser_portfolio_2011}. An increasing
number of research communities are becoming aware of Algorithm Selection
techniques and the potential benefits for their domain.

Related research explores how Algorithm Selection techniques can be used in the
construction of software
\cite{balasubramaniam_automated_2012,hoos_programming_2012}. This is not just
the application in a new problem domain, but the deployment of techniques in a
new context that has the potential for much higher performance improvements.
While at the moment Algorithm Selection is somewhat of a specialised subject,
the integration of relevant techniques into mainstream programming languages and
software development systems will stimulate further research in this direction.

\section{Summary}\label{sec:conclusion}

Over the years, there have been many approaches to solving the Algorithm
Selection Problem. Especially in Artificial Intelligence and for combinatorial
search problems, researchers have recognised that using Algorithm Selection
techniques can provide significant performance improvements with relatively
little effort. Most of the time, the approaches involve some kind of Machine
Learning that attempts to learn the relation between problems and the
performance of algorithms automatically. This is not a surprise, as the
relationship between an algorithm and its performance is often complex and hard
to describe formally. In many cases, even the designer of an algorithm does not
have a general model of its performance.

Despite the theoretical difficulty of Algorithm Selection, dozens of systems
have demonstrated that it can be done in practice with great success. In some
sense, this mirrors achievements in other areas of Artificial Intelligence.
Satisfiability is formally a problem that cannot be solved efficiently, yet
researchers have come up with ways of solving very large instances of
satisfiability problems with very few resources. Similarly, some Algorithm
Selection systems have come very close to always choosing the best algorithm.

This survey presented an overview of the Algorithm Selection research that has
been done to date with a focus on combinatorial search problems. A
categorisation of the different approaches with respect to fundamental criteria
that determine Algorithm Selection systems in practice was introduced. This
categorisation abstracts from many of the low level details and additional
considerations that are presented in most publications to give a clear view of
the underlying principles. We furthermore gave details of the many different
ways that can be used to tackle Algorithm Selection and the many techniques that
have been used to solve it in practice.

On a high level, the approaches surveyed here can be summarised as follows.
\begin{itemize}
\item Algorithms are chosen from portfolios, which can be statically
    constructed or dynamically augmented with newly constructed algorithms as
    problems are being solved. Portfolios can be engineered such that the
    algorithms in it complement each other (i.e.\ are as diverse as possible),
    by automatically tuning algorithms on a set of training problems or by using
    a set of algorithms from the literature or competitions. Dynamic portfolios
    can be composed of algorithmic building blocks that are combined into
    complete algorithms by the selection system. Compared to tuning the
    parameters of algorithms, the added difficulty is that not all combinations
    of building blocks may be valid.
\item A single algorithm can be selected from a portfolio to solve a problem to
    completion or a set of larger size can be selected that is run in parallel
    or according to a schedule. Another approach is to select a single algorithm
    to start with and then decide if and when to switch to another algorithm.
    Some approaches always select the entire portfolio and vary the resource
    allocation to the algorithms.
\item Algorithm Selection can happen offline, without any interaction with the
    Algorithm Selection system after solving starts, or online. Some approaches
    monitor the performance of the selected algorithm and take action if it does
    not conform to the expectations or some other criteria. Others repeat the
    selection process at specific points during the search (e.g.\ every node in
    the search tree), skew a computed schedule towards the best performers or
    decide whether to restart stochastic algorithms.
\item Performance can be modelled and predicted either for a portfolio as a
    whole (i.e.\ the prediction is the best algorithm) or for each algorithm
    independently (i.e.\ the prediction is the performance). A few approaches
    use hierarchical models that make a series of predictions to facilitate
    selection. Some publications make secondary predictions (e.g.\ the quality
    of a solution) that are taken into account when selecting the most suitable
    algorithm, while others make predictions that the desired output is derived
    from instead of predicting it directly. The performance models are usually
    learned automatically using Machine Learning, but a few approaches use
    hand-crafted models and rules. Models can be learned from separate training
    data or incrementally while a problem is being solved.
\item Learning and using performance models is facilitated by features of the
    algorithms, problems or runtime environment. Features can be
    domain-independent or specific to a particular set of problems. Similarly,
    features can be computed by inspecting the problem before solving or while
    it is being solved. The use of feature selection techniques that
    automatically determine the most important and relevant features is quite
    common.
\end{itemize}

Given the amount of relevant literature, it is infeasible to discuss every
approach in detail. The scope of this survey is necessarily limited to
the detailed description of high-level details and a summary overview of
low-level traits. Work in related areas that is not immediately relevant to
Algorithm Selection for combinatorial search problems has been pointed to, but
cannot be explored in more detail.

A tabular summary of the literature organised according to the criteria
introduced here can be found at
\url{http://4c.ucc.ie/~larsko/assurvey/}.

\acks

Ian Miguel and Ian Gent provided valuable feedback that helped shape this paper.
We also thank the anonymous reviewers of a previous version of this paper whose
detailed comments helped to greatly improve it. This work was supported by an
EPSRC doctoral prize.

\bibliographystyle{theapa}
\bibliography{survey}

\begin{thebibliography}{}

\bibitem[\protect\BCAY{Aha}{Aha}{1992}]{aha_generalizing_1992}
Aha, D.~W. \BBOP1992\BBCP.
\newblock \BBOQ Generalizing from case studies: A case study\BBCQ\
\newblock In {\Bem Proceedings of the 9th International Workshop on Machine
  Learning}, \BPGS\ 1--10, San Francisco, {CA}, {USA}. Morgan Kaufmann
  Publishers Inc.

\bibitem[\protect\BCAY{Allen\ \BBA\ Minton}{Allen\ \BBA\
  Minton}{1996}]{allen_selecting_1996}
Allen, J.~A.\BBACOMMA\  \BBA\ Minton, S. \BBOP1996\BBCP.
\newblock \BBOQ Selecting the right heuristic algorithm: Runtime performance
  predictors\BBCQ\
\newblock In {\Bem The 11th Biennial Conference of the Canadian Society for
  Computational Studies of Intelligence}, \BPGS\ 41--53. {Springer-Verlag}.

\bibitem[\protect\BCAY{Ansel, Chan, Wong, Olszewski, Zhao, Edelman,\ \BBA\
  Amarasinghe}{Ansel et~al.}{2009}]{ansel_petabricks_2009}
Ansel, J., Chan, C., Wong, Y.~L., Olszewski, M., Zhao, Q., Edelman, A., \BBA\
  Amarasinghe, S. \BBOP2009\BBCP.
\newblock \BBOQ {PetaBricks:} a language and compiler for algorithmic
  choice\BBCQ\
\newblock {\Bem {SIGPLAN} Not.}, {\Bem 44\/}(6), 38--49.

\bibitem[\protect\BCAY{Ans{\'o}tegui, Sellmann,\ \BBA\ Tierney}{Ans{\'o}tegui
  et~al.}{2009}]{ansotegui_gender-based_2009}
Ans{\'o}tegui, C., Sellmann, M., \BBA\ Tierney, K. \BBOP2009\BBCP.
\newblock \BBOQ A {Gender-Based} genetic algorithm for the automatic
  configuration of algorithms\BBCQ\
\newblock In {\Bem {CP}}, \BPGS\ 142--157.

\bibitem[\protect\BCAY{Arbelaez, Hamadi,\ \BBA\ Sebag}{Arbelaez
  et~al.}{2009}]{arbelaez_online_2009}
Arbelaez, A., Hamadi, Y., \BBA\ Sebag, M. \BBOP2009\BBCP.
\newblock \BBOQ Online heuristic selection in constraint programming\BBCQ\
\newblock In {\Bem Symposium on Combinatorial Search}.

\bibitem[\protect\BCAY{Armstrong, Christen, {McCreath},\ \BBA\
  Rendell}{Armstrong et~al.}{2006}]{armstrong_dynamic_2006}
Armstrong, W., Christen, P., {McCreath}, E., \BBA\ Rendell, A.~P.
  \BBOP2006\BBCP.
\newblock \BBOQ Dynamic algorithm selection using reinforcement learning\BBCQ\
\newblock In {\Bem International Workshop on Integrating {AI} and Data Mining},
  \BPGS\ 18--25.

\bibitem[\protect\BCAY{Balasubramaniam, Gent, Jefferson, Kotthoff, Miguel,\
  \BBA\ Nightingale}{Balasubramaniam
  et~al.}{2012}]{balasubramaniam_automated_2012}
Balasubramaniam, D., Gent, I.~P., Jefferson, C., Kotthoff, L., Miguel, I.,
  \BBA\ Nightingale, P. \BBOP2012\BBCP.
\newblock \BBOQ An automated approach to generating efficient constraint
  solvers\BBCQ\
\newblock In {\Bem 34th International Conference on Software Engineering},
  \BPGS\ 661--671.

\bibitem[\protect\BCAY{Bauer\ \BBA\ Kohavi}{Bauer\ \BBA\
  Kohavi}{1999}]{bauer_empirical_1999}
Bauer, E.\BBACOMMA\  \BBA\ Kohavi, R. \BBOP1999\BBCP.
\newblock \BBOQ An empirical comparison of voting classification algorithms:
  Bagging, boosting, and variants\BBCQ\
\newblock {\Bem Machine Learning}, {\Bem 36\/}(1-2), 105--139.

\bibitem[\protect\BCAY{Beck\ \BBA\ Fox}{Beck\ \BBA\
  Fox}{2000}]{beck_dynamic_2000}
Beck, J.~C.\BBACOMMA\  \BBA\ Fox, M.~S. \BBOP2000\BBCP.
\newblock \BBOQ Dynamic problem structure analysis as a basis for
  constraint-directed scheduling heuristics\BBCQ\
\newblock {\Bem Artificial Intelligence}, {\Bem 117\/}(1), 31--81.

\bibitem[\protect\BCAY{Beck\ \BBA\ Freuder}{Beck\ \BBA\
  Freuder}{2004}]{beck_simple_2004}
Beck, J.~C.\BBACOMMA\  \BBA\ Freuder, E.~C. \BBOP2004\BBCP.
\newblock \BBOQ Simple rules for low-knowledge algorithm selection\BBCQ\
\newblock In {\Bem {CPAIOR}}, \BPGS\ 50--64. Springer.

\bibitem[\protect\BCAY{Bhowmick, Eijkhout, Freund, Fuentes,\ \BBA\
  Keyes}{Bhowmick et~al.}{2006}]{bhowmick_application_2006}
Bhowmick, S., Eijkhout, V., Freund, Y., Fuentes, E., \BBA\ Keyes, D.
  \BBOP2006\BBCP.
\newblock \BBOQ Application of machine learning in selecting sparse linear
  solvers\BBCQ\
\newblock \BTR, Columbia University.

\bibitem[\protect\BCAY{Bhowmick, Toth,\ \BBA\ Raghavan}{Bhowmick
  et~al.}{2009}]{bhowmick_towards_2009}
Bhowmick, S., Toth, B., \BBA\ Raghavan, P. \BBOP2009\BBCP.
\newblock \BBOQ Towards {Low-Cost}, {High-Accuracy} classifiers for linear
  solver selection\BBCQ\
\newblock In {\Bem Proceedings of the 9th International Conference on
  Computational Science}, {ICCS} '09, \BPGS\ 463--472, Berlin, Heidelberg.
  {Springer-Verlag}.

\bibitem[\protect\BCAY{Borrett\ \BBA\ Tsang}{Borrett\ \BBA\
  Tsang}{2001}]{borrett_context_2001}
Borrett, J.~E.\BBACOMMA\  \BBA\ Tsang, E. P.~K. \BBOP2001\BBCP.
\newblock \BBOQ A context for constraint satisfaction problem formulation
  selection\BBCQ\
\newblock {\Bem Constraints}, {\Bem 6\/}(4), 299--327.

\bibitem[\protect\BCAY{Borrett, Tsang,\ \BBA\ Walsh}{Borrett
  et~al.}{1996}]{borrett_adaptive_1996}
Borrett, J.~E., Tsang, E. P.~K., \BBA\ Walsh, N.~R. \BBOP1996\BBCP.
\newblock \BBOQ Adaptive constraint satisfaction: The quickest first
  principle\BBCQ\
\newblock In {\Bem {ECAI}}, \BPGS\ 160--164.

\bibitem[\protect\BCAY{Bougeret, Dutot, Goldman, Ngoko,\ \BBA\
  Trystram}{Bougeret et~al.}{2009}]{bougeret_combining_2009}
Bougeret, M., Dutot, P., Goldman, A., Ngoko, Y., \BBA\ Trystram, D.
  \BBOP2009\BBCP.
\newblock \BBOQ Combining multiple heuristics on discrete resources\BBCQ\
\newblock In {\Bem {IEEE} International Symposium on Parallel \& Distributed
  Processing}, \BPGS\ 1--8, Washington, {DC}, {USA}. {IEEE} Computer Society.

\bibitem[\protect\BCAY{Brazdil\ \BBA\ Soares}{Brazdil\ \BBA\
  Soares}{2000}]{brazdil_comparison_2000}
Brazdil, P.\BBACOMMA\  \BBA\ Soares, C. \BBOP2000\BBCP.
\newblock \BBOQ A comparison of ranking methods for classification algorithm
  selection\BBCQ\
\newblock In {\Bem Proceedings of the 11th European Conference on Machine
  Learning}, {ECML} '00, \BPGS\ 63--74, London, {UK}. {Springer-Verlag}.

\bibitem[\protect\BCAY{Breiman}{Breiman}{1996}]{breiman_bagging_1996}
Breiman, L. \BBOP1996\BBCP.
\newblock \BBOQ Bagging predictors\BBCQ\
\newblock {\Bem Mach. Learn.}, {\Bem 24\/}(2), 123--140.

\bibitem[\protect\BCAY{Brewer}{Brewer}{1995}]{brewer_high-level_1995}
Brewer, E.~A. \BBOP1995\BBCP.
\newblock \BBOQ High-level optimization via automated statistical
  modeling\BBCQ\
\newblock In {\Bem Proceedings of the 5th {ACM} {SIGPLAN} Symposium on
  Principles and Practice of Parallel Programming}, {PPOPP} '95, \BPGS\ 80--91,
  New York, {NY}, {USA}. {ACM}.

\bibitem[\protect\BCAY{Brodley}{Brodley}{1993}]{brodley_automatic_1993}
Brodley, C.~E. \BBOP1993\BBCP.
\newblock \BBOQ Addressing the selective superiority problem: Automatic
  {Algorithm/Model} class selection\BBCQ\
\newblock In {\Bem {ICML}}, \BPGS\ 17--24.

\bibitem[\protect\BCAY{Cahill}{Cahill}{1994}]{cahill_knowledge-based_1994}
Cahill, E. \BBOP1994\BBCP.
\newblock \BBOQ Knowledge-based algorithm construction for real-world
  engineering {PDEs}\BBCQ\
\newblock {\Bem Mathematics and Computers in Simulation}, {\Bem 36\/}(4-6),
  389--400.

\bibitem[\protect\BCAY{Carbonell, Etzioni, Gil, Joseph, Knoblock, Minton,\
  \BBA\ Veloso}{Carbonell et~al.}{1991}]{carbonell_prodigy_1991}
Carbonell, J., Etzioni, O., Gil, Y., Joseph, R., Knoblock, C., Minton, S.,
  \BBA\ Veloso, M. \BBOP1991\BBCP.
\newblock \BBOQ {PRODIGY:} an integrated architecture for planning and
  learning\BBCQ\
\newblock {\Bem {SIGART} Bull.}, {\Bem 2}, 51--55.

\bibitem[\protect\BCAY{Carchrae}{Carchrae}{2009}]{carchrae_low_2009}
Carchrae, T. \BBOP2009\BBCP.
\newblock {\Bem Low Knowledge Algorithm Control for {Constraint-Based}
  Scheduling}.
\newblock Ph.D.\ thesis, National University of Ireland.

\bibitem[\protect\BCAY{Carchrae\ \BBA\ Beck}{Carchrae\ \BBA\
  Beck}{2004}]{carchrae_low-knowledge_2004}
Carchrae, T.\BBACOMMA\  \BBA\ Beck, J.~C. \BBOP2004\BBCP.
\newblock \BBOQ {Low-Knowledge} algorithm control\BBCQ\
\newblock In {\Bem {AAAI}}, \BPGS\ 49--54.

\bibitem[\protect\BCAY{Carchrae\ \BBA\ Beck}{Carchrae\ \BBA\
  Beck}{2005}]{carchrae_applying_2005}
Carchrae, T.\BBACOMMA\  \BBA\ Beck, J.~C. \BBOP2005\BBCP.
\newblock \BBOQ Applying machine learning to {Low-Knowledge} control of
  optimization algorithms\BBCQ\
\newblock {\Bem Computational Intelligence}, {\Bem 21\/}(4), 372--387.

\bibitem[\protect\BCAY{Caseau, Laburthe,\ \BBA\ Silverstein}{Caseau
  et~al.}{1999}]{caseau_meta-heuristic_1999}
Caseau, Y., Laburthe, F., \BBA\ Silverstein, G. \BBOP1999\BBCP.
\newblock \BBOQ A {Meta-Heuristic} factory for vehicle routing problems\BBCQ\
\newblock In {\Bem Proceedings of the 5th International Conference on
  Principles and Practice of Constraint Programming}, \BPGS\ 144--158, London,
  {UK}. {Springer-Verlag}.

\bibitem[\protect\BCAY{Cheeseman, Kanefsky,\ \BBA\ Taylor}{Cheeseman
  et~al.}{1991}]{cheeseman_where_1991}
Cheeseman, P., Kanefsky, B., \BBA\ Taylor, W.~M. \BBOP1991\BBCP.
\newblock \BBOQ Where the really hard problems are\BBCQ\
\newblock In {\Bem 12th International Joint Conference on Artificial
  Intelligence}, \BPGS\ 331--337, San Francisco, {CA}, {USA}. Morgan Kaufmann
  Publishers Inc.

\bibitem[\protect\BCAY{Cicirello\ \BBA\ Smith}{Cicirello\ \BBA\
  Smith}{2005}]{cicirello_max_2005}
Cicirello, V.~A.\BBACOMMA\  \BBA\ Smith, S.~F. \BBOP2005\BBCP.
\newblock \BBOQ The max k-armed bandit: A new model of exploration applied to
  search heuristic selection\BBCQ\
\newblock In {\Bem Proceedings of the 20th National Conference on Artificial
  Intelligence}, \BPGS\ 1355--1361. {AAAI} Press.

\bibitem[\protect\BCAY{Cook\ \BBA\ Varnell}{Cook\ \BBA\
  Varnell}{1997}]{cook_maximizing_1997}
Cook, D.~J.\BBACOMMA\  \BBA\ Varnell, R.~C. \BBOP1997\BBCP.
\newblock \BBOQ Maximizing the benefits of parallel search using machine
  learning\BBCQ\
\newblock In {\Bem Proceedings of the 14th National Conference on Artificial
  Intelligence}, \BPGS\ 559--564. {AAAI} Press.

\bibitem[\protect\BCAY{Demmel, Dongarra, Eijkhout, Fuentes, Petitet, Vuduc,
  Whaley,\ \BBA\ Yelick}{Demmel et~al.}{2005}]{demmel_self-adapting_2005}
Demmel, J., Dongarra, J., Eijkhout, V., Fuentes, E., Petitet, A., Vuduc, R.,
  Whaley, R.~C., \BBA\ Yelick, K. \BBOP2005\BBCP.
\newblock \BBOQ {Self-Adapting} linear algebra algorithms and software\BBCQ\
\newblock {\Bem Proceedings of the {IEEE}}, {\Bem 93\/}(2), 293--312.

\bibitem[\protect\BCAY{Dietterich}{Dietterich}{2000}]{dietterich_ensemble_2000}
Dietterich, T.~G. \BBOP2000\BBCP.
\newblock \BBOQ Ensemble methods in machine learning\BBCQ\
\newblock In {\Bem Proceedings of the 1st International Workshop on Multiple
  Classifier Systems}, \lowercase{\BVOL}\ 1857 of {\Bem Lecture Notes In
  Computer Science}, \BPGS\ 1--15. {Springer-Verlag}.

\bibitem[\protect\BCAY{Domingos}{Domingos}{1998}]{domingos_how_1998}
Domingos, P. \BBOP1998\BBCP.
\newblock \BBOQ How to get a free lunch: A simple cost model for machine
  learning applications\BBCQ\
\newblock In {\Bem {AAAI98/ICML98} Workshop on the Methodology of Applying
  Machine Learning}, \BPGS\ 1--7. {AAAI} Press.

\bibitem[\protect\BCAY{Domshlak, Karpas,\ \BBA\ Markovitch}{Domshlak
  et~al.}{2010}]{domshlak_max_2010}
Domshlak, C., Karpas, E., \BBA\ Markovitch, S. \BBOP2010\BBCP.
\newblock \BBOQ To max or not to max: Online learning for speeding up optimal
  planning\BBCQ\
\newblock In {\Bem {AAAI}}.

\bibitem[\protect\BCAY{Elsayed\ \BBA\ Michel}{Elsayed\ \BBA\
  Michel}{2010}]{elsayed_synthesis_2010}
Elsayed, S. A.~M.\BBACOMMA\  \BBA\ Michel, L. \BBOP2010\BBCP.
\newblock \BBOQ Synthesis of search algorithms from high-level {CP}
  models\BBCQ\
\newblock In {\Bem Proceedings of the 9th International Workshop on Constraint
  Modelling and Reformulation}.

\bibitem[\protect\BCAY{Elsayed\ \BBA\ Michel}{Elsayed\ \BBA\
  Michel}{2011}]{elsayed_synthesis_2011}
Elsayed, S. A.~M.\BBACOMMA\  \BBA\ Michel, L. \BBOP2011\BBCP.
\newblock \BBOQ Synthesis of search algorithms from high-level {CP}
  models\BBCQ\
\newblock In {\Bem 17th International Conference on Principles and Practice of
  Constraint Programming}, \BPGS\ 256--270, Berlin, Heidelberg.
  {Springer-Verlag}.

\bibitem[\protect\BCAY{Epstein\ \BBA\ Freuder}{Epstein\ \BBA\
  Freuder}{2001}]{epstein_collaborative_2001}
Epstein, S.~L.\BBACOMMA\  \BBA\ Freuder, E.~C. \BBOP2001\BBCP.
\newblock \BBOQ Collaborative learning for constraint solving\BBCQ\
\newblock In {\Bem Proceedings of the 7th International Conference on
  Principles and Practice of Constraint Programming}, \BPGS\ 46--60, London,
  {UK}. {Springer-Verlag}.

\bibitem[\protect\BCAY{Epstein, Freuder, Wallace, Morozov,\ \BBA\
  Samuels}{Epstein et~al.}{2002}]{epstein_adaptive_2002}
Epstein, S.~L., Freuder, E.~C., Wallace, R., Morozov, A., \BBA\ Samuels, B.
  \BBOP2002\BBCP.
\newblock \BBOQ The adaptive constraint engine\BBCQ\
\newblock In {\Bem Principles and Practice of Constraint Programming}, \BPGS\
  525--540. Springer.

\bibitem[\protect\BCAY{Ewald}{Ewald}{2010}]{ewald_automatic_2010}
Ewald, R. \BBOP2010\BBCP.
\newblock {\Bem Automatic Algorithm Selection for Complex Simulation Problems}.
\newblock Ph.D.\ thesis, University of Rostock.

\bibitem[\protect\BCAY{Ewald, Schulz,\ \BBA\ Uhrmacher}{Ewald
  et~al.}{2010}]{ewald_selecting_2010}
Ewald, R., Schulz, R., \BBA\ Uhrmacher, A.~M. \BBOP2010\BBCP.
\newblock \BBOQ Selecting simulation algorithm portfolios by genetic
  algorithms\BBCQ\
\newblock In {\Bem {IEEE} Workshop on Principles of Advanced and Distributed
  Simulation}, {PADS} '10, \BPGS\ 1--9, Washington, {DC}, {USA}. {IEEE}
  Computer Society.

\bibitem[\protect\BCAY{Fink}{Fink}{1997}]{fink_statistical_1997}
Fink, E. \BBOP1997\BBCP.
\newblock \BBOQ Statistical selection among {Problem-Solving} methods\BBCQ\
\newblock \BTR\ {CMU-CS-97-101}, Carnegie Mellon University.

\bibitem[\protect\BCAY{Fink}{Fink}{1998}]{fink_how_1998}
Fink, E. \BBOP1998\BBCP.
\newblock \BBOQ How to solve it automatically: Selection among
  {Problem-Solving} methods\BBCQ\
\newblock In {\Bem Proceedings of the 4th International Conference on
  Artificial Intelligence Planning Systems}, \BPGS\ 128--136. {AAAI} Press.

\bibitem[\protect\BCAY{Fukunaga}{Fukunaga}{2000}]{fukunaga_genetic_2000}
Fukunaga, A.~S. \BBOP2000\BBCP.
\newblock \BBOQ Genetic algorithm portfolios\BBCQ\
\newblock In {\Bem {IEEE} Congress on Evolutionary Computation},
  \lowercase{\BVOL}~2, \BPGS\ 1304--1311.

\bibitem[\protect\BCAY{Fukunaga}{Fukunaga}{2002}]{fukunaga_automated_2002}
Fukunaga, A.~S. \BBOP2002\BBCP.
\newblock \BBOQ Automated discovery of composite {SAT} variable-selection
  heuristics\BBCQ\
\newblock In {\Bem 18th National Conference on Artificial Intelligence}, \BPGS\
  641--648, Menlo Park, {CA}, {USA}. American Association for Artificial
  Intelligence.

\bibitem[\protect\BCAY{Fukunaga}{Fukunaga}{2008}]{fukunaga_automated_2008}
Fukunaga, A.~S. \BBOP2008\BBCP.
\newblock \BBOQ Automated discovery of local search heuristics for
  satisfiability testing\BBCQ\
\newblock {\Bem Evol. Comput.}, {\Bem 16}, 31--61.

\bibitem[\protect\BCAY{Gagliolo}{Gagliolo}{2010}]{gagliolo_online_2010}
Gagliolo, M. \BBOP2010\BBCP.
\newblock {\Bem Online Dynamic Algorithm Portfolios -- Minimizing the
  computational cost of problem solving}.
\newblock Ph.D.\ thesis, University of Lugano.

\bibitem[\protect\BCAY{Gagliolo\ \BBA\ Schmidhuber}{Gagliolo\ \BBA\
  Schmidhuber}{2005}]{gagliolo_neural_2005}
Gagliolo, M.\BBACOMMA\  \BBA\ Schmidhuber, J. \BBOP2005\BBCP.
\newblock \BBOQ A neural network model for {Inter-Problem} adaptive online time
  allocation\BBCQ\
\newblock In {\Bem 15th International Conference on Artificial Neural Networks:
  Formal Models and Their Applications}, \BPGS\ 7--12. Springer.

\bibitem[\protect\BCAY{Gagliolo\ \BBA\ Schmidhuber}{Gagliolo\ \BBA\
  Schmidhuber}{2006a}]{gagliolo_impact_2006}
Gagliolo, M.\BBACOMMA\  \BBA\ Schmidhuber, J. \BBOP2006a\BBCP.
\newblock \BBOQ Impact of censored sampling on the performance of restart
  strategies\BBCQ\
\newblock In {\Bem {CP}}, \BPGS\ 167--181.

\bibitem[\protect\BCAY{Gagliolo\ \BBA\ Schmidhuber}{Gagliolo\ \BBA\
  Schmidhuber}{2006b}]{gagliolo_learning_2006}
Gagliolo, M.\BBACOMMA\  \BBA\ Schmidhuber, J. \BBOP2006b\BBCP.
\newblock \BBOQ Learning dynamic algorithm portfolios\BBCQ\
\newblock {\Bem Ann. Math. Artif. Intell.}, {\Bem 47\/}(3-4), 295--328.

\bibitem[\protect\BCAY{Gagliolo\ \BBA\ Schmidhuber}{Gagliolo\ \BBA\
  Schmidhuber}{2008}]{gagliolo_towards_2008}
Gagliolo, M.\BBACOMMA\  \BBA\ Schmidhuber, J. \BBOP2008\BBCP.
\newblock \BBOQ Towards distributed algorithm portfolios\BBCQ\
\newblock In {\Bem International Symposium on Distributed Computing and
  Artificial Intelligence, Advances in Soft Computing}. Springer.

\bibitem[\protect\BCAY{Gagliolo\ \BBA\ Schmidhuber}{Gagliolo\ \BBA\
  Schmidhuber}{2011}]{gagliolo_algorithm_2011}
Gagliolo, M.\BBACOMMA\  \BBA\ Schmidhuber, J. \BBOP2011\BBCP.
\newblock \BBOQ Algorithm portfolio selection as a bandit problem with
  unbounded losses\BBCQ\
\newblock {\Bem Annals of Mathematics and Artificial Intelligence}, {\Bem
  61\/}(2), 49--86.

\bibitem[\protect\BCAY{Gagliolo, Zhumatiy,\ \BBA\ Schmidhuber}{Gagliolo
  et~al.}{2004}]{gagliolo_adaptive_2004}
Gagliolo, M., Zhumatiy, V., \BBA\ Schmidhuber, J. \BBOP2004\BBCP.
\newblock \BBOQ Adaptive online time allocation to search algorithms\BBCQ\
\newblock In {\Bem {ECML}}, \BPGS\ 134--143. Springer.

\bibitem[\protect\BCAY{Garrido\ \BBA\ Riff}{Garrido\ \BBA\
  Riff}{2010}]{garrido_dvrp_2010}
Garrido, P.\BBACOMMA\  \BBA\ Riff, M. \BBOP2010\BBCP.
\newblock \BBOQ {DVRP:} a hard dynamic combinatorial optimisation problem
  tackled by an evolutionary hyper-heuristic\BBCQ\
\newblock {\Bem Journal of Heuristics}, {\Bem 16}, 795--834.

\bibitem[\protect\BCAY{Gebruers, Guerri, Hnich,\ \BBA\ Milano}{Gebruers
  et~al.}{2004}]{gebruers_making_2004}
Gebruers, C., Guerri, A., Hnich, B., \BBA\ Milano, M. \BBOP2004\BBCP.
\newblock \BBOQ Making choices using structure at the instance level within a
  case based reasoning framework\BBCQ\
\newblock In {\Bem {CPAIOR}}, \BPGS\ 380--386.

\bibitem[\protect\BCAY{Gebruers, Hnich, Bridge,\ \BBA\ Freuder}{Gebruers
  et~al.}{2005}]{gebruers_using_2005}
Gebruers, C., Hnich, B., Bridge, D., \BBA\ Freuder, E. \BBOP2005\BBCP.
\newblock \BBOQ Using {CBR} to select solution strategies in constraint
  programming\BBCQ\
\newblock In {\Bem Proc. of {ICCBR-05}}, \BPGS\ 222--236.

\bibitem[\protect\BCAY{Gebser, Kaminski, Kaufmann, Schaub, Schneider,\ \BBA\
  Ziller}{Gebser et~al.}{2011}]{gebser_portfolio_2011}
Gebser, M., Kaminski, R., Kaufmann, B., Schaub, T., Schneider, M.~T., \BBA\
  Ziller, S. \BBOP2011\BBCP.
\newblock \BBOQ A portfolio solver for answer set programming: preliminary
  report\BBCQ\
\newblock In {\Bem 11th International Conference on Logic Programming and
  Nonmonotonic Reasoning}, \BPGS\ 352--357, Berlin, Heidelberg.
  {Springer-Verlag}.

\bibitem[\protect\BCAY{Gent, Jefferson, Kotthoff, Miguel, Moore, Nightingale,\
  \BBA\ Petrie}{Gent et~al.}{2010a}]{gent_learning_2010}
Gent, I., Jefferson, C., Kotthoff, L., Miguel, I., Moore, N., Nightingale, P.,
  \BBA\ Petrie, K. \BBOP2010a\BBCP.
\newblock \BBOQ Learning when to use lazy learning in constraint solving\BBCQ\
\newblock In {\Bem 19th European Conference on Artificial Intelligence}, \BPGS\
  873--878.

\bibitem[\protect\BCAY{Gent, Kotthoff, Miguel,\ \BBA\ Nightingale}{Gent
  et~al.}{2010b}]{gent_machine_2010}
Gent, I., Kotthoff, L., Miguel, I., \BBA\ Nightingale, P. \BBOP2010b\BBCP.
\newblock \BBOQ Machine learning for constraint solver design -– a case study
  for the alldifferent constraint\BBCQ\
\newblock In {\Bem 3rd Workshop on Techniques for implementing Constraint
  Programming Systems {(TRICS)}}, \BPGS\ 13--25.

\bibitem[\protect\BCAY{Gerevini, Saetti,\ \BBA\ Vallati}{Gerevini
  et~al.}{2009}]{gerevini_automatically_2009}
Gerevini, A.~E., Saetti, A., \BBA\ Vallati, M. \BBOP2009\BBCP.
\newblock \BBOQ An automatically configurable portfolio-based planner with
  macro-actions: {PbP}\BBCQ\
\newblock In {\Bem Proceedings of the 19th International Conference on
  Automated Planning and Scheduling}, \BPGS\ 350--353.

\bibitem[\protect\BCAY{Gomes\ \BBA\ Selman}{Gomes\ \BBA\
  Selman}{1997a}]{gomes_algorithm_1997}
Gomes, C.~P.\BBACOMMA\  \BBA\ Selman, B. \BBOP1997a\BBCP.
\newblock \BBOQ Algorithm portfolio design: Theory vs. practice\BBCQ\
\newblock In {\Bem {UAI}}, \BPGS\ 190--197.

\bibitem[\protect\BCAY{Gomes\ \BBA\ Selman}{Gomes\ \BBA\
  Selman}{1997b}]{gomes_practical_1997}
Gomes, C.~P.\BBACOMMA\  \BBA\ Selman, B. \BBOP1997b\BBCP.
\newblock \BBOQ Practical aspects of algorithm portfolio design\BBCQ\
\newblock In {\Bem Proc. of 3rd {ILOG} International Users Meeting}.

\bibitem[\protect\BCAY{Gomes\ \BBA\ Selman}{Gomes\ \BBA\
  Selman}{2001}]{gomes_algorithm_2001}
Gomes, C.~P.\BBACOMMA\  \BBA\ Selman, B. \BBOP2001\BBCP.
\newblock \BBOQ Algorithm portfolios\BBCQ\
\newblock {\Bem Artificial Intelligence}, {\Bem 126\/}(1-2), 43--62.

\bibitem[\protect\BCAY{Gratch\ \BBA\ {DeJong}}{Gratch\ \BBA\
  {DeJong}}{1992}]{gratch_composer_1992}
Gratch, J.\BBACOMMA\  \BBA\ {DeJong}, G. \BBOP1992\BBCP.
\newblock \BBOQ {COMPOSER:} a probabilistic solution to the utility problem in
  {Speed-Up} learning\BBCQ\
\newblock In {\Bem {AAAI}}, \BPGS\ 235--240.

\bibitem[\protect\BCAY{Guerri\ \BBA\ Milano}{Guerri\ \BBA\
  Milano}{2004}]{guerri_learning_2004}
Guerri, A.\BBACOMMA\  \BBA\ Milano, M. \BBOP2004\BBCP.
\newblock \BBOQ Learning techniques for automatic algorithm portfolio
  selection\BBCQ\
\newblock In {\Bem {ECAI}}, \BPGS\ 475--479.

\bibitem[\protect\BCAY{Guo}{Guo}{2003}]{guo_algorithm_2003}
Guo, H. \BBOP2003\BBCP.
\newblock {\Bem Algorithm Selection for Sorting and Probabilistic Inference: A
  Machine {Learning-Based} Approach}.
\newblock Ph.D.\ thesis, Kansas State University.

\bibitem[\protect\BCAY{Guo\ \BBA\ Hsu}{Guo\ \BBA\
  Hsu}{2004}]{guo_learning-based_2004}
Guo, H.\BBACOMMA\  \BBA\ Hsu, W.~H. \BBOP2004\BBCP.
\newblock \BBOQ A {Learning-Based} algorithm selection meta-reasoner for the
  {Real-Time} {MPE} problem\BBCQ\
\newblock In {\Bem Australian Conference on Artificial Intelligence}, \BPGS\
  307--318.

\bibitem[\protect\BCAY{Haim\ \BBA\ Walsh}{Haim\ \BBA\
  Walsh}{2009}]{haim_restart_2009}
Haim, S.\BBACOMMA\  \BBA\ Walsh, T. \BBOP2009\BBCP.
\newblock \BBOQ Restart strategy selection using machine learning
  techniques\BBCQ\
\newblock In {\Bem Proceedings of the 12th International Conference on Theory
  and Applications of Satisfiability Testing}, \BPGS\ 312--325, Berlin,
  Heidelberg. {Springer-Verlag}.

\bibitem[\protect\BCAY{Hogg, Huberman,\ \BBA\ Williams}{Hogg
  et~al.}{1996}]{hogg_phase_1996}
Hogg, T., Huberman, B.~A., \BBA\ Williams, C.~P. \BBOP1996\BBCP.
\newblock \BBOQ Phase transitions and the search problem\BBCQ\
\newblock {\Bem Artif. Intell.}, {\Bem 81\/}(1-2), 1--15.

\bibitem[\protect\BCAY{Hong\ \BBA\ Page}{Hong\ \BBA\
  Page}{2004}]{hong_groups_2004}
Hong, L.\BBACOMMA\  \BBA\ Page, S.~E. \BBOP2004\BBCP.
\newblock \BBOQ Groups of diverse problem solvers can outperform groups of
  high-ability problem solvers\BBCQ\
\newblock {\Bem Proceedings of the National Academy of Sciences of the United
  States of America}, {\Bem 101\/}(46), 16385--16389.

\bibitem[\protect\BCAY{Hoos}{Hoos}{2012}]{hoos_programming_2012}
Hoos, H.~H. \BBOP2012\BBCP.
\newblock \BBOQ Programming by optimization\BBCQ\
\newblock {\Bem Commun. {ACM}}, {\Bem 55\/}(2), 70--80.

\bibitem[\protect\BCAY{Horvitz, Ruan, Gomes, Kautz, Selman,\ \BBA\
  Chickering}{Horvitz et~al.}{2001}]{horvitz_bayesian_2001}
Horvitz, E., Ruan, Y., Gomes, C.~P., Kautz, H.~A., Selman, B., \BBA\
  Chickering, D.~M. \BBOP2001\BBCP.
\newblock \BBOQ A bayesian approach to tackling hard computational
  problems\BBCQ\
\newblock In {\Bem Proceedings of the 17th Conference in Uncertainty in
  Artificial Intelligence}, \BPGS\ 235--244, San Francisco, {CA}, {USA}. Morgan
  Kaufmann Publishers Inc.

\bibitem[\protect\BCAY{Hough\ \BBA\ Williams}{Hough\ \BBA\
  Williams}{2006}]{hough_modern_2006}
Hough, P.~D.\BBACOMMA\  \BBA\ Williams, P.~J. \BBOP2006\BBCP.
\newblock \BBOQ Modern machine learning for automatic optimization algorithm
  selection\BBCQ\
\newblock In {\Bem Proceedings of the {INFORMS} Artificial Intelligence and
  Data Mining Workshop}.

\bibitem[\protect\BCAY{Howe, Dahlman, Hansen, Scheetz,\ \BBA\ von
  Mayrhauser}{Howe et~al.}{1999}]{howe_exploiting_1999}
Howe, A.~E., Dahlman, E., Hansen, C., Scheetz, M., \BBA\ von Mayrhauser, A.
  \BBOP1999\BBCP.
\newblock \BBOQ Exploiting competitive planner performance\BBCQ\
\newblock In {\Bem Proceedings of the 5th European Conference on Planning},
  \BPGS\ 62--72. Springer.

\bibitem[\protect\BCAY{Huberman, Lukose,\ \BBA\ Hogg}{Huberman
  et~al.}{1997}]{huberman_economics_1997}
Huberman, B.~A., Lukose, R.~M., \BBA\ Hogg, T. \BBOP1997\BBCP.
\newblock \BBOQ An economics approach to hard computational problems\BBCQ\
\newblock {\Bem Science}, {\Bem 275\/}(5296), 51--54.

\bibitem[\protect\BCAY{Hutter}{Hutter}{2009}]{hutter_automated_2009}
Hutter, F. \BBOP2009\BBCP.
\newblock {\Bem Automated Configuration of Algorithms for Solving Hard
  Computational Problems}.
\newblock Ph.D.\ thesis, University of British Columbia, Department of Computer
  Science, Vancouver, Canada.

\bibitem[\protect\BCAY{Hutter, Hamadi, Hoos,\ \BBA\ {Leyton-Brown}}{Hutter
  et~al.}{2006}]{hutter_performance_2006}
Hutter, F., Hamadi, Y., Hoos, H.~H., \BBA\ {Leyton-Brown}, K. \BBOP2006\BBCP.
\newblock \BBOQ Performance prediction and automated tuning of randomized and
  parametric algorithms\BBCQ\
\newblock In {\Bem CP}, \BPGS\ 213--228.

\bibitem[\protect\BCAY{Hutter, Hoos,\ \BBA\ {Leyton-Brown}}{Hutter
  et~al.}{2012}]{hutter_parallel_2012}
Hutter, F., Hoos, H.~H., \BBA\ {Leyton-Brown}, K. \BBOP2012\BBCP.
\newblock \BBOQ Parallel algorithm configuration\BBCQ\
\newblock In {\Bem Proc. of {LION-6}}.

\bibitem[\protect\BCAY{Hutter, Hoos, {Leyton-Brown},\ \BBA\ St\"utzle}{Hutter
  et~al.}{2009}]{hutter_paramils_2009}
Hutter, F., Hoos, H.~H., {Leyton-Brown}, K., \BBA\ St\"utzle, T.
  \BBOP2009\BBCP.
\newblock \BBOQ {ParamILS:} an automatic algorithm configuration
  framework\BBCQ\
\newblock {\Bem J. Artif. Int. Res.}, {\Bem 36\/}(1), 267--306.

\bibitem[\protect\BCAY{Hutter, Hoos,\ \BBA\ St\"utzle}{Hutter
  et~al.}{2007}]{hutter_automatic_2007}
Hutter, F., Hoos, H.~H., \BBA\ St\"utzle, T. \BBOP2007\BBCP.
\newblock \BBOQ Automatic algorithm configuration based on local search\BBCQ\
\newblock In {\Bem Proceedings of the 22nd National Conference on Artificial
  Intelligence}, \BPGS\ 1152--1157. {AAAI} Press.

\bibitem[\protect\BCAY{Kadioglu, Malitsky, Sabharwal, Samulowitz,\ \BBA\
  Sellmann}{Kadioglu et~al.}{2011}]{kadioglu_algorithm_2011}
Kadioglu, S., Malitsky, Y., Sabharwal, A., Samulowitz, H., \BBA\ Sellmann, M.
  \BBOP2011\BBCP.
\newblock \BBOQ Algorithm selection and scheduling\BBCQ\
\newblock In {\Bem 17th International Conference on Principles and Practice of
  Constraint Programming}, \BPGS\ 454--469.

\bibitem[\protect\BCAY{Kadioglu, Malitsky, Sellmann,\ \BBA\ Tierney}{Kadioglu
  et~al.}{2010}]{kadioglu_isac_2010}
Kadioglu, S., Malitsky, Y., Sellmann, M., \BBA\ Tierney, K. \BBOP2010\BBCP.
\newblock \BBOQ {ISAC} – {Instance-Specific} algorithm configuration\BBCQ\
\newblock In {\Bem 19th European Conference on Artificial Intelligence}, \BPGS\
  751--756. {IOS} Press.

\bibitem[\protect\BCAY{Kamel, Enright,\ \BBA\ Ma}{Kamel
  et~al.}{1993}]{kamel_odexpert_1993}
Kamel, M.~S., Enright, W.~H., \BBA\ Ma, K.~S. \BBOP1993\BBCP.
\newblock \BBOQ {ODEXPERT:} an expert system to select numerical solvers for
  initial value {ODE} systems\BBCQ\
\newblock {\Bem {ACM} Trans. Math. Softw.}, {\Bem 19\/}(1), 44--62.

\bibitem[\protect\BCAY{Kotthoff}{Kotthoff}{2012a}]{kotthoff_hybrid_2012}
Kotthoff, L. \BBOP2012a\BBCP.
\newblock \BBOQ Hybrid regression-classification models for algorithm
  selection\BBCQ\
\newblock In {\Bem 20th European Conference on Artificial Intelligence}, \BPGS\
  480--485.

\bibitem[\protect\BCAY{Kotthoff}{Kotthoff}{2012b}]{kotthoff_algorithm_2012}
Kotthoff, L. \BBOP2012b\BBCP.
\newblock {\Bem On Algorithm Selection, with an Application to Combinatorial
  Search Problems}.
\newblock Ph.D.\ thesis, University of St Andrews.

\bibitem[\protect\BCAY{Kotthoff, Gent,\ \BBA\ Miguel}{Kotthoff
  et~al.}{2012}]{kotthoff_evaluation_2012}
Kotthoff, L., Gent, I.~P., \BBA\ Miguel, I. \BBOP2012\BBCP.
\newblock \BBOQ An evaluation of machine learning in algorithm selection for
  search problems\BBCQ\
\newblock {\Bem {AI} Communications}, {\Bem 25\/}(3), 257--270.

\bibitem[\protect\BCAY{Kotthoff, Miguel,\ \BBA\ Nightingale}{Kotthoff
  et~al.}{2010}]{kotthoff_ensemble_2010}
Kotthoff, L., Miguel, I., \BBA\ Nightingale, P. \BBOP2010\BBCP.
\newblock \BBOQ Ensemble classification for constraint solver
  configuration\BBCQ\
\newblock In {\Bem 16th International Conference on Principles and Practices of
  Constraint Programming}, \BPGS\ 321--329.

\bibitem[\protect\BCAY{Kroer\ \BBA\ Malitsky}{Kroer\ \BBA\
  Malitsky}{2011}]{kroer_feature_2011}
Kroer, C.\BBACOMMA\  \BBA\ Malitsky, Y. \BBOP2011\BBCP.
\newblock \BBOQ Feature filtering for {Instance-Specific} algorithm
  configuration\BBCQ\
\newblock In {\Bem Proceedings of the 23rd International Conference on Tools
  with Artificial Intelligence}.

\bibitem[\protect\BCAY{Kuefler\ \BBA\ Chen}{Kuefler\ \BBA\
  Chen}{2008}]{kuefler_using_2008}
Kuefler, E.\BBACOMMA\  \BBA\ Chen, T. \BBOP2008\BBCP.
\newblock \BBOQ On using reinforcement learning to solve sparse linear
  systems\BBCQ\
\newblock In {\Bem Proceedings of the 8th International Conference on
  Computational Science}, {ICCS} '08, \BPGS\ 955--964, Berlin, Heidelberg.
  {Springer-Verlag}.

\bibitem[\protect\BCAY{Lagoudakis\ \BBA\ Littman}{Lagoudakis\ \BBA\
  Littman}{2000}]{lagoudakis_algorithm_2000}
Lagoudakis, M.~G.\BBACOMMA\  \BBA\ Littman, M.~L. \BBOP2000\BBCP.
\newblock \BBOQ Algorithm selection using reinforcement learning\BBCQ\
\newblock In {\Bem Proceedings of the 17th International Conference on Machine
  Learning}, \BPGS\ 511--518, San Francisco, {CA}, {USA}. Morgan Kaufmann
  Publishers Inc.

\bibitem[\protect\BCAY{Lagoudakis\ \BBA\ Littman}{Lagoudakis\ \BBA\
  Littman}{2001}]{lagoudakis_learning_2001}
Lagoudakis, M.~G.\BBACOMMA\  \BBA\ Littman, M.~L. \BBOP2001\BBCP.
\newblock \BBOQ Learning to select branching rules in the {DPLL} procedure for
  satisfiability\BBCQ\
\newblock In {\Bem {LICS/SAT}}, \BPGS\ 344--359.

\bibitem[\protect\BCAY{Langley}{Langley}{1983a}]{langley_learning_1983}
Langley, P. \BBOP1983a\BBCP.
\newblock \BBOQ Learning effective search heuristics\BBCQ\
\newblock In {\Bem {IJCAI}}, \BPGS\ 419--421.

\bibitem[\protect\BCAY{Langley}{Langley}{1983b}]{langley_learningd_1983}
Langley, P. \BBOP1983b\BBCP.
\newblock \BBOQ Learning search strategies through discrimination.\BBCQ\
\newblock {\Bem International Journal of {Man-Machine} Studies}, 513--541.

\bibitem[\protect\BCAY{Leite, Brazdil, Vanschoren,\ \BBA\ Queiros}{Leite
  et~al.}{2010}]{leite_using_2010}
Leite, R., Brazdil, P., Vanschoren, J., \BBA\ Queiros, F. \BBOP2010\BBCP.
\newblock \BBOQ Using active testing and {Meta-Level} information for selection
  of classification algorithms\BBCQ\
\newblock In {\Bem 3rd {PlanLearn} Workshop}.

\bibitem[\protect\BCAY{{Leyton-Brown}, Nudelman,\ \BBA\ Shoham}{{Leyton-Brown}
  et~al.}{2002}]{leyton-brown_learning_2002}
{Leyton-Brown}, K., Nudelman, E., \BBA\ Shoham, Y. \BBOP2002\BBCP.
\newblock \BBOQ Learning the empirical hardness of optimization problems: The
  case of combinatorial auctions\BBCQ\
\newblock In {\Bem Proceedings of the 8th International Conference on
  Principles and Practice of Constraint Programming}, \BPGS\ 556--572, London,
  {UK}. {Springer-Verlag}.

\bibitem[\protect\BCAY{{Leyton-Brown}, Nudelman,\ \BBA\ Shoham}{{Leyton-Brown}
  et~al.}{2009}]{leyton-brown_empirical_2009}
{Leyton-Brown}, K., Nudelman, E., \BBA\ Shoham, Y. \BBOP2009\BBCP.
\newblock \BBOQ Empirical hardness models: Methodology and a case study on
  combinatorial auctions\BBCQ\
\newblock {\Bem J. {ACM}}, {\Bem 56}, 22:1--22:52.

\bibitem[\protect\BCAY{Little, Gebruers, Bridge,\ \BBA\ Freuder}{Little
  et~al.}{2002}]{little_capturing_2002}
Little, J., Gebruers, C., Bridge, D., \BBA\ Freuder, E. \BBOP2002\BBCP.
\newblock \BBOQ Capturing constraint programming experience: A {Case-Based}
  approach\BBCQ\
\newblock In {\Bem Modref}.

\bibitem[\protect\BCAY{Lobjois\ \BBA\ Lema\^itre}{Lobjois\ \BBA\
  Lema\^itre}{1998}]{lobjois_branch_1998}
Lobjois, L.\BBACOMMA\  \BBA\ Lema\^itre, M. \BBOP1998\BBCP.
\newblock \BBOQ Branch and bound algorithm selection by performance
  prediction\BBCQ\
\newblock In {\Bem Proceedings of the 15th National/10th Conference on
  Artificial {Intelligence/Innovative} Applications of Artificial
  Intelligence}, \BPGS\ 353--358, Menlo Park, {CA}, {USA}. American Association
  for Artificial Intelligence.

\bibitem[\protect\BCAY{Malitsky}{Malitsky}{2012}]{malitsky_thesis_2012}
Malitsky, Y. \BBOP2012\BBCP.
\newblock {\Bem {Instance-Specific} Algorithm Configuration}.
\newblock Ph.D.\ thesis, Brown University.

\bibitem[\protect\BCAY{Malitsky, Sabharwal, Samulowitz,\ \BBA\
  Sellmann}{Malitsky et~al.}{2011}]{malitsky_non-model-based_2011}
Malitsky, Y., Sabharwal, A., Samulowitz, H., \BBA\ Sellmann, M. \BBOP2011\BBCP.
\newblock \BBOQ Non-model-based algorithm portfolios for {SAT}\BBCQ\
\newblock In {\Bem Theory and Applications of Satisfiability Testing {(SAT)}},
  \BPGS\ 369--370.

\bibitem[\protect\BCAY{Minton}{Minton}{1993a}]{minton_analytic_1993}
Minton, S. \BBOP1993a\BBCP.
\newblock \BBOQ An analytic learning system for specializing heuristics\BBCQ\
\newblock In {\Bem {IJCAI'93:} Proceedings of the 13th International Joint
  Conference on Artifical Intelligence}, \BPGS\ 922--928, San Francisco, {CA},
  {USA}. Morgan Kaufmann Publishers Inc.

\bibitem[\protect\BCAY{Minton}{Minton}{1993b}]{minton_integrating_1993}
Minton, S. \BBOP1993b\BBCP.
\newblock \BBOQ Integrating heuristics for constraint satisfaction problems: A
  case study\BBCQ\
\newblock In {\Bem {AAAI:} Proceedings of the 11th National Conference on
  Artificial Intelligence}, \BPGS\ 120--126.

\bibitem[\protect\BCAY{Minton}{Minton}{1996}]{minton_automatically_1996}
Minton, S. \BBOP1996\BBCP.
\newblock \BBOQ Automatically configuring constraint satisfaction programs: A
  case study\BBCQ\
\newblock {\Bem Constraints}, {\Bem 1}, 7--43.

\bibitem[\protect\BCAY{Nareyek}{Nareyek}{2001}]{nareyek_choosing_2001}
Nareyek, A. \BBOP2001\BBCP.
\newblock \BBOQ Choosing search heuristics by {Non-Stationary} reinforcement
  learning\BBCQ\
\newblock In {\Bem Metaheuristics: Computer {Decision-Making}}, \BPGS\
  523--544. Kluwer Academic Publishers.

\bibitem[\protect\BCAY{Nikoli\'c, Mari\'c,\ \BBA\ Jani\v{c}i\'c}{Nikoli\'c
  et~al.}{2009}]{nikoli_instance-based_2009}
Nikoli\'c, M., Mari\'c, F., \BBA\ Jani\v{c}i\'c, P. \BBOP2009\BBCP.
\newblock \BBOQ {Instance-Based} selection of policies for {SAT} solvers\BBCQ\
\newblock In {\Bem Proceedings of the 12th International Conference on Theory
  and Applications of Satisfiability Testing}, {SAT} '09, \BPGS\ 326--340,
  Berlin, Heidelberg. {Springer-Verlag}.

\bibitem[\protect\BCAY{Nudelman, {Leyton-Brown}, Hoos, Devkar,\ \BBA\
  Shoham}{Nudelman et~al.}{2004}]{nudelman_understanding_2004}
Nudelman, E., {Leyton-Brown}, K., Hoos, H.~H., Devkar, A., \BBA\ Shoham, Y.
  \BBOP2004\BBCP.
\newblock \BBOQ Understanding random {SAT:} beyond the {Clauses-to-Variables}
  ratio\BBCQ\
\newblock In Wallace, M.\BED, {\Bem Principles and Practice of Constraint
  Programming – {CP} 2004}, \lowercase{\BVOL}\ 3258 of {\Bem Lecture Notes in
  Computer Science}, \BPGS\ 438--452. Springer Berlin / Heidelberg.

\bibitem[\protect\BCAY{{O'Mahony}, Hebrard, Holland, Nugent,\ \BBA\
  {O'Sullivan}}{{O'Mahony} et~al.}{2008}]{omahony_using_2008}
{O'Mahony}, E., Hebrard, E., Holland, A., Nugent, C., \BBA\ {O'Sullivan}, B.
  \BBOP2008\BBCP.
\newblock \BBOQ Using case-based reasoning in an algorithm portfolio for
  constraint solving\BBCQ\
\newblock In {\Bem Proceedings of the 19th Irish Conference on Artificial
  Intelligence and Cognitive Science}.

\bibitem[\protect\BCAY{Opitz\ \BBA\ Maclin}{Opitz\ \BBA\
  Maclin}{1999}]{opitz_popular_1999}
Opitz, D.\BBACOMMA\  \BBA\ Maclin, R. \BBOP1999\BBCP.
\newblock \BBOQ Popular ensemble methods: An empirical study\BBCQ\
\newblock {\Bem Journal of Artificial Intelligence Research}, {\Bem 11},
  169--198.

\bibitem[\protect\BCAY{Petrik}{Petrik}{2005}]{petrik_statistically_2005}
Petrik, M. \BBOP2005\BBCP.
\newblock \BBOQ Statistically optimal combination of algorithms\BBCQ\
\newblock In {\Bem Local Proceedings of {SOFSEM} 2005}.

\bibitem[\protect\BCAY{Petrik\ \BBA\ Zilberstein}{Petrik\ \BBA\
  Zilberstein}{2006}]{petrik_learning_2006}
Petrik, M.\BBACOMMA\  \BBA\ Zilberstein, S. \BBOP2006\BBCP.
\newblock \BBOQ Learning parallel portfolios of algorithms\BBCQ\
\newblock {\Bem Annals of Mathematics and Artificial Intelligence}, {\Bem
  48\/}(1-2), 85--106.

\bibitem[\protect\BCAY{Petrovic\ \BBA\ Qu}{Petrovic\ \BBA\
  Qu}{2002}]{petrovic_case-based_2002}
Petrovic, S.\BBACOMMA\  \BBA\ Qu, R. \BBOP2002\BBCP.
\newblock \BBOQ {Case-Based} reasoning as a heuristic selector in
  {Hyper-Heuristic} for course timetabling problems\BBCQ\
\newblock In {\Bem {KES}}, \BPGS\ 336--340.

\bibitem[\protect\BCAY{Pfahringer, Bensusan,\ \BBA\
  {Giraud-Carrier}}{Pfahringer et~al.}{2000}]{pfahringer_meta-learning_2000}
Pfahringer, B., Bensusan, H., \BBA\ {Giraud-Carrier}, C.~G. \BBOP2000\BBCP.
\newblock \BBOQ {Meta-Learning} by landmarking various learning
  algorithms\BBCQ\
\newblock In {\Bem 17th International Conference on Machine Learning}, {ICML}
  '00, \BPGS\ 743--750, San Francisco, {CA}, {USA}. Morgan Kaufmann Publishers
  Inc.

\bibitem[\protect\BCAY{Pulina\ \BBA\ Tacchella}{Pulina\ \BBA\
  Tacchella}{2007}]{pulina_multi-engine_2007}
Pulina, L.\BBACOMMA\  \BBA\ Tacchella, A. \BBOP2007\BBCP.
\newblock \BBOQ A multi-engine solver for quantified boolean formulas\BBCQ\
\newblock In {\Bem Proceedings of the 13th International Conference on
  Principles and Practice of Constraint Programming}, {CP'07}, \BPGS\ 574--589,
  Berlin, Heidelberg. {Springer-Verlag}.

\bibitem[\protect\BCAY{Pulina\ \BBA\ Tacchella}{Pulina\ \BBA\
  Tacchella}{2009}]{pulina_self-adaptive_2009}
Pulina, L.\BBACOMMA\  \BBA\ Tacchella, A. \BBOP2009\BBCP.
\newblock \BBOQ A self-adaptive multi-engine solver for quantified boolean
  formulas\BBCQ\
\newblock {\Bem Constraints}, {\Bem 14\/}(1), 80--116.

\bibitem[\protect\BCAY{Rao, Gordon,\ \BBA\ Spears}{Rao
  et~al.}{1995}]{rao_for_1995}
Rao, R.~B., Gordon, D., \BBA\ Spears, W. \BBOP1995\BBCP.
\newblock \BBOQ For every generalization action, is there really an equal and
  opposite reaction? {Analysis} of the conservation law for generalization
  performance\BBCQ\
\newblock In {\Bem Proceedings of the 12th International Conference on Machine
  Learning}, \BPGS\ 471--479. Morgan Kaufmann.

\bibitem[\protect\BCAY{Rice}{Rice}{1976}]{rice_algorithm_1976}
Rice, J.~R. \BBOP1976\BBCP.
\newblock \BBOQ The algorithm selection problem\BBCQ\
\newblock {\Bem Advances in Computers}, {\Bem 15}, 65--118.

\bibitem[\protect\BCAY{Rice\ \BBA\ Ramakrishnan}{Rice\ \BBA\
  Ramakrishnan}{1999}]{rice_how_1999}
Rice, J.~R.\BBACOMMA\  \BBA\ Ramakrishnan, N. \BBOP1999\BBCP.
\newblock \BBOQ How to get a free lunch (at no cost)\BBCQ\
\newblock \BTR\ 99-014, Purdue University.

\bibitem[\protect\BCAY{Roberts\ \BBA\ Howe}{Roberts\ \BBA\
  Howe}{2006}]{roberts_directing_2006}
Roberts, M.\BBACOMMA\  \BBA\ Howe, A.~E. \BBOP2006\BBCP.
\newblock \BBOQ Directing a portfolio with learning\BBCQ\
\newblock In {\Bem {AAAI} 2006 Workshop on Learning for Search}.

\bibitem[\protect\BCAY{Roberts\ \BBA\ Howe}{Roberts\ \BBA\
  Howe}{2007}]{roberts_learned_2007}
Roberts, M.\BBACOMMA\  \BBA\ Howe, A.~E. \BBOP2007\BBCP.
\newblock \BBOQ Learned models of performance for many planners\BBCQ\
\newblock In {\Bem {ICAPS} 2007 Workshop {AI} Planning and Learning}.

\bibitem[\protect\BCAY{Roberts, Howe, Wilson,\ \BBA\ {desJardins}}{Roberts
  et~al.}{2008}]{roberts_what_2008}
Roberts, M., Howe, A.~E., Wilson, B., \BBA\ {desJardins}, M. \BBOP2008\BBCP.
\newblock \BBOQ What makes planners predictable?\BBCQ\
\newblock In {\Bem {ICAPS}}, \BPGS\ 288--295.

\bibitem[\protect\BCAY{Sakkout, Wallace,\ \BBA\ Richards}{Sakkout
  et~al.}{1996}]{sakkout_instance_1996}
Sakkout, H.~E., Wallace, M.~G., \BBA\ Richards, E.~B. \BBOP1996\BBCP.
\newblock \BBOQ An instance of adaptive constraint propagation\BBCQ\
\newblock In {\Bem Proc. of {CP96}}, \BPGS\ 164--178. Springer Verlag.

\bibitem[\protect\BCAY{Samulowitz\ \BBA\ Memisevic}{Samulowitz\ \BBA\
  Memisevic}{2007}]{samulowitz_learning_2007}
Samulowitz, H.\BBACOMMA\  \BBA\ Memisevic, R. \BBOP2007\BBCP.
\newblock \BBOQ Learning to solve {QBF}\BBCQ\
\newblock In {\Bem Proceedings of the 22nd National Conference on Artificial
  Intelligence}, \BPGS\ 255--260. {AAAI} Press.

\bibitem[\protect\BCAY{Sayag, Fine,\ \BBA\ Mansour}{Sayag
  et~al.}{2006}]{sayag_combining_2006}
Sayag, T., Fine, S., \BBA\ Mansour, Y. \BBOP2006\BBCP.
\newblock \BBOQ Combining multiple heuristics\BBCQ\
\newblock In {\Bem {STACS}}, \lowercase{\BVOL}\ 3884, \BPGS\ 242--253, Berlin,
  Heidelberg. Springer.

\bibitem[\protect\BCAY{Schapire}{Schapire}{1990}]{schapire_strength_1990}
Schapire, R.~E. \BBOP1990\BBCP.
\newblock \BBOQ The strength of weak learnability\BBCQ\
\newblock {\Bem Machine Learning}, {\Bem 5\/}(2), 197--227.

\bibitem[\protect\BCAY{Sillito}{Sillito}{2000}]{sillito_improvements_2000}
Sillito, J. \BBOP2000\BBCP.
\newblock \BBOQ Improvements to and estimating the cost of solving constraint
  satisfaction problems\BBCQ\
\newblock Master's thesis, University of Alberta.

\bibitem[\protect\BCAY{Silverthorn\ \BBA\ Miikkulainen}{Silverthorn\ \BBA\
  Miikkulainen}{2010}]{silverthorn_latent_2010}
Silverthorn, B.\BBACOMMA\  \BBA\ Miikkulainen, R. \BBOP2010\BBCP.
\newblock \BBOQ Latent class models for algorithm portfolio methods\BBCQ\
\newblock In {\Bem Proceedings of the 24th {AAAI} Conference on Artificial
  Intelligence}.

\bibitem[\protect\BCAY{Smith\ \BBA\ Setliff}{Smith\ \BBA\
  Setliff}{1992}]{smith_knowledge-based_1992}
Smith, T.~E.\BBACOMMA\  \BBA\ Setliff, D.~E. \BBOP1992\BBCP.
\newblock \BBOQ Knowledge-based constraint-driven software synthesis\BBCQ\
\newblock In {\Bem {Knowledge-Based} Software Engineering Conference}, \BPGS\
  18--27.

\bibitem[\protect\BCAY{{Smith-Miles}\ \BBA\ Lopes}{{Smith-Miles}\ \BBA\
  Lopes}{2012}]{smith-miles_measuring_2012}
{Smith-Miles}, K.\BBACOMMA\  \BBA\ Lopes, L. \BBOP2012\BBCP.
\newblock \BBOQ Measuring instance difficulty for combinatorial optimization
  problems\BBCQ\
\newblock {\Bem Comput. Oper. Res.}, {\Bem 39\/}(5), 875--889.

\bibitem[\protect\BCAY{{Smith-Miles}}{{Smith-Miles}}{2008a}]{smith-miles_cross-disciplinary_2009}
{Smith-Miles}, K.~A. \BBOP2008a\BBCP.
\newblock \BBOQ Cross-disciplinary perspectives on meta-learning for algorithm
  selection\BBCQ\
\newblock {\Bem {ACM} Comput. Surv.}, {\Bem 41}, 6:1--6:25.

\bibitem[\protect\BCAY{{Smith-Miles}}{{Smith-Miles}}{2008b}]{smith-miles_towards_2008}
{Smith-Miles}, K.~A. \BBOP2008b\BBCP.
\newblock \BBOQ Towards insightful algorithm selection for optimisation using
  {Meta-Learning} concepts\BBCQ\
\newblock In {\Bem {IEEE} International Joint Conference on Neural Networks},
  \BPGS\ 4118--4124.

\bibitem[\protect\BCAY{Soares, Brazdil,\ \BBA\ Kuba}{Soares
  et~al.}{2004}]{soares_meta-learning_2004}
Soares, C., Brazdil, P.~B., \BBA\ Kuba, P. \BBOP2004\BBCP.
\newblock \BBOQ A {Meta-Learning} method to select the kernel width in support
  vector regression\BBCQ\
\newblock {\Bem Mach. Learn.}, {\Bem 54\/}(3), 195--209.

\bibitem[\protect\BCAY{Stamatatos\ \BBA\ Stergiou}{Stamatatos\ \BBA\
  Stergiou}{2009}]{stamatatos_learning_2009}
Stamatatos, E.\BBACOMMA\  \BBA\ Stergiou, K. \BBOP2009\BBCP.
\newblock \BBOQ Learning how to propagate using random probing\BBCQ\
\newblock In {\Bem Proceedings of the 6th International Conference on
  Integration of {AI} and {OR} Techniques in Constraint Programming for
  Combinatorial Optimization Problems}, \BPGS\ 263--278, Berlin, Heidelberg.
  {Springer-Verlag}.

\bibitem[\protect\BCAY{Stergiou}{Stergiou}{2009}]{stergiou_heuristics_2009}
Stergiou, K. \BBOP2009\BBCP.
\newblock \BBOQ Heuristics for dynamically adapting propagation in constraint
  satisfaction problems\BBCQ\
\newblock {\Bem {AI} Commun.}, {\Bem 22\/}(3), 125--141.

\bibitem[\protect\BCAY{Stern, Samulowitz, Herbrich, Graepel, Pulina,\ \BBA\
  Tacchella}{Stern et~al.}{2010}]{stern_collaborative_2010}
Stern, D.~H., Samulowitz, H., Herbrich, R., Graepel, T., Pulina, L., \BBA\
  Tacchella, A. \BBOP2010\BBCP.
\newblock \BBOQ Collaborative expert portfolio management\BBCQ\
\newblock In {\Bem {AAAI}}, \BPGS\ 179--184.

\bibitem[\protect\BCAY{Streeter}{Streeter}{2007}]{streeter_using_2007}
Streeter, M.~J. \BBOP2007\BBCP.
\newblock {\Bem Using Online Algorithms to Solve {NP-Hard} Problems More
  Efficiently in Practice}.
\newblock Ph.D.\ thesis, Carnegie Mellon University.

\bibitem[\protect\BCAY{Streeter, Golovin,\ \BBA\ Smith}{Streeter
  et~al.}{2007a}]{streeter_combining_2007}
Streeter, M.~J., Golovin, D., \BBA\ Smith, S.~F. \BBOP2007a\BBCP.
\newblock \BBOQ Combining multiple heuristics online\BBCQ\
\newblock In {\Bem Proceedings of the 22nd National Conference on Artificial
  Intelligence}, \BPGS\ 1197--1203. {AAAI} Press.

\bibitem[\protect\BCAY{Streeter, Golovin,\ \BBA\ Smith}{Streeter
  et~al.}{2007b}]{streeter_restart_2007}
Streeter, M.~J., Golovin, D., \BBA\ Smith, S.~F. \BBOP2007b\BBCP.
\newblock \BBOQ Restart schedules for ensembles of problem instances\BBCQ\
\newblock In {\Bem Proceedings of the 22nd National Conference on Artificial
  Intelligence}, \BPGS\ 1204--1210. {AAAI} Press.

\bibitem[\protect\BCAY{Streeter\ \BBA\ Smith}{Streeter\ \BBA\
  Smith}{2008}]{streeter_new_2008}
Streeter, M.~J.\BBACOMMA\  \BBA\ Smith, S.~F. \BBOP2008\BBCP.
\newblock \BBOQ New techniques for algorithm portfolio design\BBCQ\
\newblock In {\Bem {UAI}}, \BPGS\ 519--527.

\bibitem[\protect\BCAY{{Terashima-Mar\'in}, Ross,\ \BBA\
  {Valenzuela-Rend\'on}}{{Terashima-Mar\'in}
  et~al.}{1999}]{terashima-marin_evolution_1999}
{Terashima-Mar\'in}, H., Ross, P., \BBA\ {Valenzuela-Rend\'on}, M.
  \BBOP1999\BBCP.
\newblock \BBOQ Evolution of constraint satisfaction strategies in examination
  timetabling\BBCQ\
\newblock In {\Bem Proceedings of the Genetic and Evolutionary Computation
  Conference}, \BPGS\ 635--642. Morgan Kaufmann.

\bibitem[\protect\BCAY{Tolpin\ \BBA\ Shimony}{Tolpin\ \BBA\
  Shimony}{2011}]{tolpin_rational_2011}
Tolpin, D.\BBACOMMA\  \BBA\ Shimony, S.~E. \BBOP2011\BBCP.
\newblock \BBOQ Rational deployment of {CSP} heuristics\BBCQ\
\newblock In {\Bem {IJCAI}}, \BPGS\ 680--686.

\bibitem[\protect\BCAY{Tsang, Borrett,\ \BBA\ Kwan}{Tsang
  et~al.}{1995}]{tsang_attempt_1995}
Tsang, E. P.~K., Borrett, J.~E., \BBA\ Kwan, A. C.~M. \BBOP1995\BBCP.
\newblock \BBOQ An attempt to map the performance of a range of algorithm and
  heuristic combinations\BBCQ\
\newblock In {\Bem Proc. of {AISB'95}}, \BPGS\ 203--216. {IOS} Press.

\bibitem[\protect\BCAY{Utgoff}{Utgoff}{1988}]{utgoff_perceptron_1988}
Utgoff, P.~E. \BBOP1988\BBCP.
\newblock \BBOQ Perceptron trees: A case study in hybrid concept
  representations\BBCQ\
\newblock In {\Bem National Conference on Artificial Intelligence}, \BPGS\
  601--606.

\bibitem[\protect\BCAY{Vassilevska, Williams,\ \BBA\ Woo}{Vassilevska
  et~al.}{2006}]{vassilevska_confronting_2006}
Vassilevska, V., Williams, R., \BBA\ Woo, S. L.~M. \BBOP2006\BBCP.
\newblock \BBOQ Confronting hardness using a hybrid approach\BBCQ\
\newblock In {\Bem Proceedings of the 17th Annual {ACM-SIAM} Symposium on
  Discrete Algorithms}, {SODA} '06, \BPGS\ 1--10, New York, {NY}, {USA}. {ACM}.

\bibitem[\protect\BCAY{Vrakas, Tsoumakas, Bassiliades,\ \BBA\ Vlahavas}{Vrakas
  et~al.}{2003}]{vrakas_learning_2003}
Vrakas, D., Tsoumakas, G., Bassiliades, N., \BBA\ Vlahavas, I. \BBOP2003\BBCP.
\newblock \BBOQ Learning rules for adaptive planning\BBCQ\
\newblock In {\Bem Proceedings of the 13th International Conference on
  Automated Planning and Scheduling}, \BPGS\ 82--91.

\bibitem[\protect\BCAY{Wang\ \BBA\ Tropper}{Wang\ \BBA\
  Tropper}{2007}]{wang_optimizing_2007}
Wang, J.\BBACOMMA\  \BBA\ Tropper, C. \BBOP2007\BBCP.
\newblock \BBOQ Optimizing time warp simulation with reinforcement learning
  techniques\BBCQ\
\newblock In {\Bem Proceedings of the 39th conference on Winter simulation},
  {WSC} '07, \BPGS\ 577--584, Piscataway, {NJ}, {USA}. {IEEE} Press.

\bibitem[\protect\BCAY{Watson}{Watson}{2003}]{watson_empirical_2003}
Watson, J. \BBOP2003\BBCP.
\newblock {\Bem Empirical modeling and analysis of local search algorithms for
  the job-shop scheduling problem}.
\newblock Ph.D.\ thesis, Colorado State University, Fort Collins, {CO}, {USA}.

\bibitem[\protect\BCAY{Weerawarana, Houstis, Rice, Joshi,\ \BBA\
  Houstis}{Weerawarana et~al.}{1996}]{weerawarana_pythia_1996}
Weerawarana, S., Houstis, E.~N., Rice, J.~R., Joshi, A., \BBA\ Houstis, C.~E.
  \BBOP1996\BBCP.
\newblock \BBOQ {PYTHIA:} a knowledge-based system to select scientific
  algorithms\BBCQ\
\newblock {\Bem {ACM} Trans. Math. Softw.}, {\Bem 22\/}(4), 447--468.

\bibitem[\protect\BCAY{Wei, Li,\ \BBA\ Zhang}{Wei
  et~al.}{2008}]{wei_switching_2008}
Wei, W., Li, C.~M., \BBA\ Zhang, H. \BBOP2008\BBCP.
\newblock \BBOQ Switching among {Non-Weighting}, clause weighting, and variable
  weighting in local search for {SAT}\BBCQ\
\newblock In {\Bem Proceedings of the 14th International Conference on
  Principles and Practice of Constraint Programming}, \BPGS\ 313--326, Berlin,
  Heidelberg. {Springer-Verlag}.

\bibitem[\protect\BCAY{Wilson, Leake,\ \BBA\ Bramley}{Wilson
  et~al.}{2000}]{wilson_case-based_2000}
Wilson, D., Leake, D., \BBA\ Bramley, R. \BBOP2000\BBCP.
\newblock \BBOQ {Case-Based} recommender components for scientific
  {Problem-Solving} environments\BBCQ\
\newblock In {\Bem Proc. of the 16th International Association for Mathematics
  and Computers in Simulation World Congress}.

\bibitem[\protect\BCAY{Wolpert}{Wolpert}{1992}]{wolpert_stacked_1992}
Wolpert, D.~H. \BBOP1992\BBCP.
\newblock \BBOQ Stacked generalization\BBCQ\
\newblock {\Bem Neural Networks}, {\Bem 5}, 241--259.

\bibitem[\protect\BCAY{Wolpert}{Wolpert}{2001}]{wolpert_supervised_2001}
Wolpert, D.~H. \BBOP2001\BBCP.
\newblock \BBOQ The supervised learning {No-Free-Lunch} theorems\BBCQ\
\newblock In {\Bem Proceedings of the 6th Online World Conference on Soft
  Computing in Industrial Applications}, \BPGS\ 25--42.

\bibitem[\protect\BCAY{Wolpert\ \BBA\ Macready}{Wolpert\ \BBA\
  Macready}{1997}]{wolpert_no_1997}
Wolpert, D.~H.\BBACOMMA\  \BBA\ Macready, W.~G. \BBOP1997\BBCP.
\newblock \BBOQ No free lunch theorems for optimization\BBCQ\
\newblock {\Bem {IEEE} Transactions on Evolutionary Computation}, {\Bem
  1\/}(1), 67--82.

\bibitem[\protect\BCAY{Wu\ \BBA\ van Beek}{Wu\ \BBA\ van
  Beek}{2007}]{wu_portfolios_2007}
Wu, H.\BBACOMMA\  \BBA\ van Beek, P. \BBOP2007\BBCP.
\newblock \BBOQ On portfolios for backtracking search in the presence of
  deadlines\BBCQ\
\newblock In {\Bem Proceedings of the 19th {IEEE} International Conference on
  Tools with Artificial Intelligence}, \BPGS\ 231--238, Washington, {DC},
  {USA}. {IEEE} Computer Society.

\bibitem[\protect\BCAY{Xu, Hoos,\ \BBA\ {Leyton-Brown}}{Xu
  et~al.}{2007}]{xu_hierarchical_2007}
Xu, L., Hoos, H.~H., \BBA\ {Leyton-Brown}, K. \BBOP2007\BBCP.
\newblock \BBOQ Hierarchical hardness models for {SAT}\BBCQ\
\newblock In {\Bem CP}, \BPGS\ 696--711.

\bibitem[\protect\BCAY{Xu, Hoos,\ \BBA\ {Leyton-Brown}}{Xu
  et~al.}{2010}]{xu_hydra_2010}
Xu, L., Hoos, H.~H., \BBA\ {Leyton-Brown}, K. \BBOP2010\BBCP.
\newblock \BBOQ Hydra: Automatically configuring algorithms for
  {Portfolio-Based} selection\BBCQ\
\newblock In {\Bem 24th Conference of the Association for the Advancement of
  Artificial Intelligence {(AAAI-10)}}, \BPGS\ 210--216.

\bibitem[\protect\BCAY{Xu, Hutter, Hoos,\ \BBA\ {Leyton-Brown}}{Xu
  et~al.}{2007}]{xu_satzilla-07_2007}
Xu, L., Hutter, F., Hoos, H.~H., \BBA\ {Leyton-Brown}, K. \BBOP2007\BBCP.
\newblock \BBOQ {SATzilla-07:} the design and analysis of an algorithm
  portfolio for {SAT}\BBCQ\
\newblock In {\Bem CP}, \BPGS\ 712--727.

\bibitem[\protect\BCAY{Xu, Hutter, Hoos,\ \BBA\ {Leyton-Brown}}{Xu
  et~al.}{2008}]{xu_satzilla_2008}
Xu, L., Hutter, F., Hoos, H.~H., \BBA\ {Leyton-Brown}, K. \BBOP2008\BBCP.
\newblock \BBOQ {SATzilla:} portfolio-based algorithm selection for {SAT}\BBCQ\
\newblock {\Bem J. Artif. Intell. Res. {(JAIR)}}, {\Bem 32}, 565--606.

\bibitem[\protect\BCAY{Xu, Hutter, Hoos,\ \BBA\ {Leyton-Brown}}{Xu
  et~al.}{2009}]{xu_satzilla2009_2009}
Xu, L., Hutter, F., Hoos, H.~H., \BBA\ {Leyton-Brown}, K. \BBOP2009\BBCP.
\newblock \BBOQ {SATzilla2009:} an automatic algorithm portfolio for
  {SAT}\BBCQ\
\newblock In {\Bem 2009 {SAT} Competition}.

\bibitem[\protect\BCAY{Xu, Hutter, Hoos,\ \BBA\ {Leyton-Brown}}{Xu
  et~al.}{2011}]{xu_hydra-mip_2011}
Xu, L., Hutter, F., Hoos, H.~H., \BBA\ {Leyton-Brown}, K. \BBOP2011\BBCP.
\newblock \BBOQ {Hydra-MIP:} automated algorithm configuration and selection
  for mixed integer programming\BBCQ\
\newblock In {\Bem {RCRA} Workshop on Experimental Evaluation of Algorithms for
  Solving Problems with Combinatorial Explosion at the International Joint
  Conference on Artificial Intelligence {(IJCAI)}}.

\bibitem[\protect\BCAY{Xu, Hutter, Hoos,\ \BBA\ {Leyton-Brown}}{Xu
  et~al.}{2012}]{xu_evaluating_2012}
Xu, L., Hutter, F., Hoos, H.~H., \BBA\ {Leyton-Brown}, K. \BBOP2012\BBCP.
\newblock \BBOQ Evaluating component solver contributions to {Portfolio-Based}
  algorithm selectors\BBCQ\
\newblock In {\Bem International Conference on Theory and Applications of
  Satisfiability Testing {(SAT'12)}}, \BPGS\ 228--241.

\bibitem[\protect\BCAY{Yu\ \BBA\ Rauchwerger}{Yu\ \BBA\
  Rauchwerger}{2006}]{yu_adaptive_2006}
Yu, H.\BBACOMMA\  \BBA\ Rauchwerger, L. \BBOP2006\BBCP.
\newblock \BBOQ An adaptive algorithm selection framework for reduction
  parallelization\BBCQ\
\newblock {\Bem {IEEE} Transactions on Parallel and Distributed Systems}, {\Bem
  17\/}(10), 1084--1096.

\bibitem[\protect\BCAY{Yu, Zhang,\ \BBA\ Rauchwerger}{Yu
  et~al.}{2004}]{yu_adaptive_2004}
Yu, H., Zhang, D., \BBA\ Rauchwerger, L. \BBOP2004\BBCP.
\newblock \BBOQ An adaptive algorithm selection framework\BBCQ\
\newblock In {\Bem Proceedings of the 13th International Conference on Parallel
  Architectures and Compilation Techniques}, \BPGS\ 278--289, Washington, {DC},
  {USA}. {IEEE} Computer Society.

\bibitem[\protect\BCAY{Yun\ \BBA\ Epstein}{Yun\ \BBA\
  Epstein}{2012}]{yun_learning_2012}
Yun, X.\BBACOMMA\  \BBA\ Epstein, S.~L. \BBOP2012\BBCP.
\newblock \BBOQ Learning algorithm portfolios for parallel execution\BBCQ\
\newblock In {\Bem Proceedings of the 6th International Conference Learning and
  Intelligent Optimisation {LION}}. Springer.

\end{thebibliography}

\end{document}